\pdfoutput=1

\documentclass[10pt,twocolumn,twoside,a4paper]{IEEEtran}
\newcommand*{\IEEEVersion}{}%

\usepackage{cite}
\usepackage{ifthen}
\usepackage{multicol}
\usepackage[utf8]{inputenc}

\usepackage[footnotesize]{caption}
\usepackage[cmex10]{amsmath}
\usepackage{amssymb}
\usepackage{amsthm}
\usepackage{amsfonts}
\usepackage{graphicx}
\usepackage{algorithm}
\usepackage{algorithmic}
\floatname{algorithm}{\footnotesize Algorithm}
\usepackage{nccmath}
\usepackage{nicefrac}
\usepackage[square,comma,numbers,sort&compress]{natbib}

\ifdefined\springerVersion
  \newboolean{publ}

  \newenvironment{bmcformat}{\begin{raggedright}\baselineskip20pt\sloppy\setboolean{publ}{false}}{\end{raggedright}\baselineskip20pt\sloppy}

  \usepackage[nolists]{endfloat}
  \DeclareDelayedFloat{algorithm}{Algorithm}
\fi

\begin{document}
\ifdefined\springerVersion
  \begin{bmcformat}
\fi

\title{Cooperative localization by dual foot-mounted inertial sensors and inter-agent ranging}

\ifdefined\IEEEVersion

\author{John-Olof~Nilsson, Dave~Zachariah, Isaac~Skog, and Peter~H\"{a}ndel
\thanks{John-Olof~Nilsson$^\ast$, Dave~Zachariah, Isaac~Skog, and Peter~H\"{a}ndel are with the Signal Processing Dept., ACCESS Linnaeus Centre, KTH Royal Institute of Technology, Osquldas v\"{a}g 10, SE-$100\,44$ Stockholm, Sweden.\protect\\E-mail: \{jnil02,davez,skog,ph\}@kth.se\protect\\\indent Parts of this work have been fonded by the Swedish Agency for Innovation Systems (VINNOVA). The authors have no connection to any company whose products are referenced in the article.}}
\fi

\ifdefined\springerVersion
\author{John-Olof~Nilsson\correspondingauthor$^1$%
         \email{John-Olof~Nilsson\correspondingauthor - jnil02@kth.se}
	\and Dave~Zachariah$^1$%
         \email{Dave~Zachariah - davez@kth.se}
	\and Isaac~Skog$^1$%
         \email{Isaac~Skog - skog@kth.se}
       and
          Peter~H\"{a}ndel$^1$%
         \email{ Peter~H\"{a}ndel - ph@kth.se}%
      }

\address{%
    \iid(1) Signal Processing Dept., ACCESS Linnaeus Centre, KTH Royal Institute of Technology, Osquldas v\"{a}g 10, SE-$100\,44$ Stockholm
}%
\fi

\maketitle

\begin{abstract}
The implementation challenges of cooperative localization by dual foot-mounted inertial sensors and inter-agent ranging are discussed and work on the subject is reviewed. System architecture and sensor fusion are identified as key challenges. A partially decentralized system architecture based on step-wise inertial navigation and step-wise dead reckoning is presented. This architecture is argued to reduce the computational cost and required communication bandwidth by around two orders of magnitude while only giving negligible information loss in comparison with a naive centralized implementation. This makes a joint global state estimation feasible for up to a platoon-sized group of agents. Furthermore, robust and low-cost sensor fusion for the considered setup, based on state space transformation and marginalization, is presented. The transformation and marginalization are used to give the necessary flexibility for presented sampling based updates for the inter-agent ranging and ranging free fusion of the two feet of an individual agent. Finally, characteristics of the suggested implementation are demonstrated with simulations and a real-time system implementation.
\end{abstract}

\ifdefined\springerVersion
\ifthenelse{\boolean{publ}}{\begin{multicols}{2}}{}
\fi

%

\section{Introduction}
High accuracy, robust, and infrastructure-free pedestrian localization is a highly desired ability for, among others, military, security personnel, and first responders. Localization together with communication are key capabilities to achieve situational awareness and to support, manage, and automatize individual's or agent group actions and interactions. See~\cite{Rantakokko2011,Fuchs2011,Pahlavan2002,Renaudin2007,Glanzer2012,Fischer2010,Hightower2001,Angermann2008} for reviews on the subject. The fundamental information sources for the localization are proprioception, exteroception, and motion models. Without infrastructure, the exteroception must be dependent on prior or acquired knowledge about the environment~\cite{Mourikis2006}. Unfortunately, in general, little or no prior knowledge of the environment is available and exploiting acquired knowledge without revisiting locations is difficult. Therefore, preferably the localization should primarily rely on proprioception and motion models. Proprioception can take place on the agent level, providing the possibility to perform dead reckoning; or on inter-agent level, providing the means to perform cooperative localization. Pedestrian dead reckoning can be implemented in a number of different ways~\cite{Harle2013}. However, foot-mounted inertial navigation, with motion models providing zero-velocity updates, constitute a unique, robust, and high accuracy pedestrian dead reckoning capability~\cite{Foxlin2005,Rantakokko2012,Godha2006,Laverne2011}. With open-source implementations~\cite{Nilsson2012,Fischer2012,instk} and several products appearing on the market~\cite{InterSense,aionav,MINT,WPI}, dead reckoning by foot-mounted inertial sensors is a readily available technology. In turn, the most straight-forward and well studied inter-agent measurement, and mean of cooperative localization, is ranging~\cite{Gezici2005,Patwari2005,Win2011,Wymeersch2009}. Also here, there are multiple (radio) ranging implementations available in the research literature~\cite{Angelis2013,Alessio2009,Moragrega2009,Karbownik2012,Cazzorla2013} and as products on the market~\cite{TimeDomain,Ensco1,nanotron}. Finally, suitable infrastructure-free communication equipment for inter-agent communication is available off-the-shelf, e.g.~\cite{Motorola,Interspiro,Thales,Zephyr}, and processing platforms are available in abundance. Together this suggests that the setup with foot-mounted inertial sensors and inter-agent ranging as illustrated in Fig.~\ref{fig:system_setup} is suitably used as a base setup for any infrastructure-free pedestrian localization system. However, despite the mature components and the in principle straight-forward combination, cooperative localization with this sensor setup remains challenging, and only a few similar systems can be found in the literature~\cite{Stromback2010,Hawkinson2012,Kloch2011,Kamisaka2012,Harris2013} 
and no off-the-shelf products are available.

\begin{figure}[t]
\centering
{\resizebox{\linewidth}{!}{\includegraphics{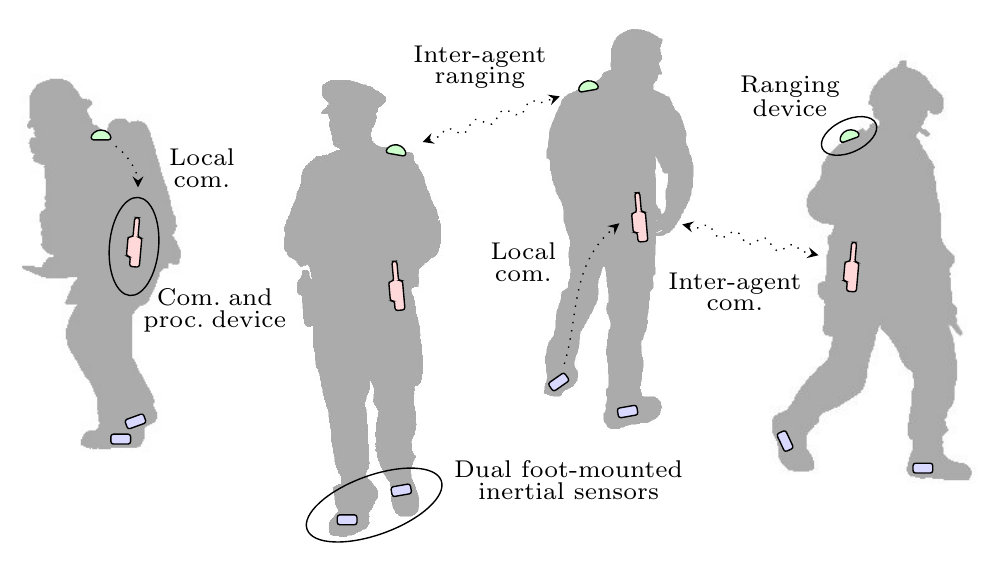}}}
\caption{Illustration of the considered localization setup. A group of agents are cooperatively localizing themselves without any infrastructure. For this purpose, each agent is equipped with dual foot-mounted inertial sensors, a ranging device, and a communication (com.) and processing (proc.) device.}\label{fig:system_setup}
\end{figure}

The challenges with the localization setup lie in the system architecture and the sensor fusion. The inter-agent ranging and lack of anchor nodes mean that some global state estimation is required with a potentially prohibitive large computational cost. The global state estimation, the distributed measurements, and the (required) high sampling rates of the inertial sensors mean that a potentially substantial communication is needed and that the system may be sensitive to incomplete or varying connectivity. The feet are poor placements for inter-agent ranging devices and preferably inertial sensors are used on both feet 
meaning that the sensors of an individual agent will not be collocated. This gives a high system state dimensionality and means that relating the sensory data from different sensors to each other is difficult and that local communication links on each agent are needed. Further, inter-agent ranging errors as well as sensor separations, often have far from white, stationary, and Gaussian characteristics. Together, this makes fusing ranging and dead reckoning in a high integrity and recursive Bayesian manner at a reasonable computational cost difficult.

Unfortunately, the mentioned challenges are inherent to the system setup. Therefore, they have to be addressed for any practical implementation. However, to our knowledge, the implementation issues have only been sparsely covered in isolation in the literature and no complete satisfactory solution has been presented. Therefore, in this article we present solutions to key challenges to the system setup and a complete localization system implementation. More specifically, the considered overall problem is tracking, i.e recursively estimating, the positions of a group of agents with the equipment setup of Fig.~\ref{fig:system_setup}. The available measurements for the tracking are inertial measurements from the dual foot-mounted inertial sensors and inter-agent range measurements. The position tracking is illustrated in Fig.~\ref{fig:localization}. The measurements will be treated as localized to the respective sensors and the necessary communication will be handled as an integral part of the overall problem. However, we will not consider specific communication technologies, but only communication constraints that naturally arise in the current scenario (low bandwidth and varying connectivity). See \cite{Adhoc,adhocnetworks,Bruno2005,Blazevic2005,Mauve2001} and references therein for treatment of related networking and communication technologies. 
Also, for brevity, the issues of initialization and time synchronization will not be considered. See~\cite{Nilsson2013a,Nilsson2010a} for the solutions used in the system implementation.

To arrive at the key challenges and the solutions, initially, in Section~\ref{sec:impl_challenges}, the implementation challenges are discussed in more detail and related work is reviewed. Following this, we address the key challenges and present a cooperative localization system implementation based on dual foot-mounted inertial sensors and inter-agent ranging. The implementation is based on a partially decentralized system architecture and statistical marginalization and sampling based measurement updates. In Section~\ref{sec:sys_arch}, the architecture is presented and argued to reduce the computational cost and required communication by around two orders of magnitude and to make the system robust to varying connectivity, while only giving negligible information loss. Thereafter, in Section~\ref{sec:sensor_fusion}, the sampling based measurement updates with required state space transformation and marginalization are presented and shown to give a robust and low computational cost sensor fusion.
Subsequently, in Section~\ref{sec:exp_results}, the characteristic of the suggested implementation is illustrated via simulations and a real-time system implementation. The cooperative localization is found to give a bounded relative position mean-square-error (MSE) and an absolute position MSE inversely proportional to the number of agents, in the worst case scenario; and a bounded position MSE, in the best case scenario. Finally, Section~\ref{sec:Conclusions} concludes the article.

\begin{figure}[t]
\centering
{\resizebox{0.98\linewidth}{!}{\includegraphics{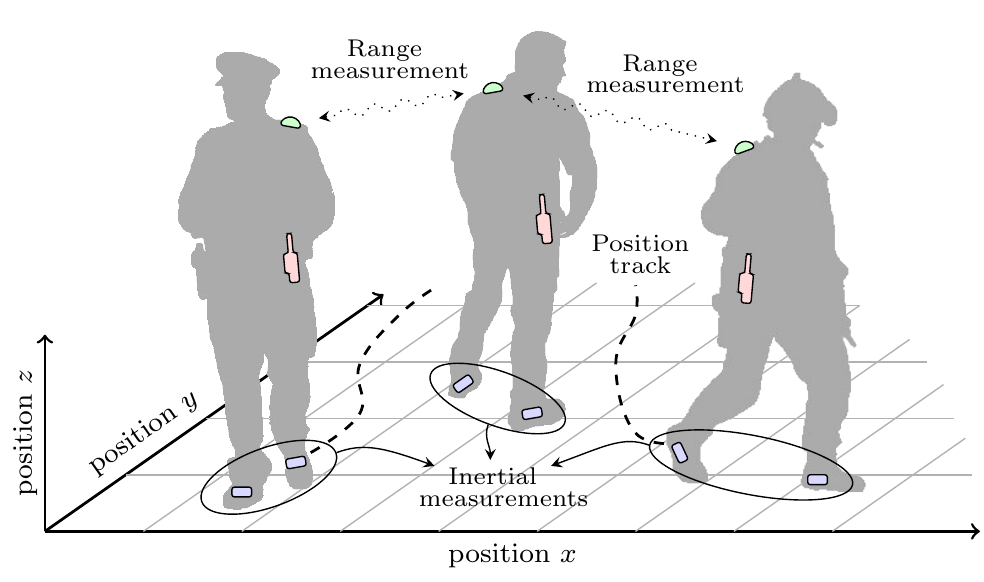}}}
\caption{Illustration of the localization estimation problem and the desired output from the localization system. The problem is to track (i.e. recursively estimate) the positions of multiple agents in three dimensions by inter-agent range measurements and inertial measurements from the foot-mounted inertial sensors.}\label{fig:localization}
\end{figure}

\section{Implementation challenges}\label{sec:impl_challenges}
The lack of anchor nodes, the distributed nature of the system, the error characteristics of the different sensors, and the non-collocated sensors of individual agents poses a number of implementation challenges for the cooperative localization. 
Broadly speaking, these challenges can be divided into those related to: designing an overall system architecture to minimize the required communication and computational cost, while making it robust to varying connectivity and retaining sufficient information about the coupling of non-collocated parts of the system; and fusing the information of different parts of the system given the constraints imposed by the system architecture and a finite computational power, while retaining performance and a high system integrity.
In the following two subsections, these two overall challenges are discussed in more detail and related previous work is reviewed.


\subsection{System architecture and state estimation}

The system architecture has a strong connection to the position/state estimation and the required communication. The range of potential system architectures and estimation solutions goes from the completely decentralized, in which each agent only estimates its own states, to the completely centralized, in which all states of all agents are estimated jointly. 

A completely decentralized architecture is often used in combination with some inherently decentralized belief-propagation estimation techniques~\cite{Stromback2010,Kschischang2001,Ihler2005}. The advantage of this is that it makes the localization scalable and robust to varying and incomplete connectivity between the agents. 
Unfortunately, the belief-propagation discards information about the coupling between agents, leading to reduced performance~\cite{Mazuelas2011,Ihler2005,Bahr2009a,Nerurkar2010}. See~\cite{Mazuelas2011} for an explicit treatment of the subject. 
Unfortunately, as will be shown in Section~\ref{sec:exp_results}, in a system with dead reckoning, inter-agent ranging, and no anchor nodes, the errors in the position estimates of different agents may become almost perfectly correlated. Consequently, discarding these couplings/correlations between agents can significantly deteriorate the localization performance and integrity. 

In contrast, with a centralized architecture and estimation,
all correlations can be considered, but instead the state dimensionality of all agents will add up. Unfortunately, due to the lack of collocation of the sensors of individual agents, the state dimensionality of individual agents will be high. Together this means computationally expensive filter updates. 
%
Further, the distributed nature of the system means that information needs to be gathered to perform the sensor fusion. Therefore, communication links are needed, both locally on each agent as well as on a group level. Inter-agent communication links are naturally wireless. However, the foot-mounting of the inertial sensors makes cabled connections impractical opting for battery powering and local wireless links for the sensors as well~\cite{Romanovas2012,Gadeke2012}. 
Unfortunately, the expensive filter updates, the wireless communication links, and the battery powering combines poorly with the required high sampling rates of the inertial sensors. With increasing number of agents, the computational cost and the required communication bandwidth will eventually become a problem.
Moreover, an agent which loses contact with the fusion center cannot, unless state statistics are continually provided, easily carry on the estimation of its own states by itself. Also, to recover from an outage when the contact is restored, a significant amount of data would have to be stored, transferred and processed.  
%

Obviously, neither of the extreme cases, the completely decentralized nor the completely centralized architectures, are acceptable. The related problems suggest that some degree of decentralization of the estimation is required to cope with the state dimensionality and communication problems. However, some global book keeping is also required to handle the information coupling. Multiple approximative and exact distributed implementations of global state estimation have been demonstrated, see~\cite{Nerurkar2010,Roumeliotis2002,Martinelli2007,Bahr2009} and references therein. However, these methods suffer from either a high computational cost or required guaranteed and high bandwidth communication, and are not adapted to the considered sensor setup with high update rates, local communication links, and lack of sensor collocation.
%
Therefore, in Section \ref{sec:sys_arch} we suggest and motivate a system architecture with partially decentralized estimation based on a division of the foot-mounted inertial navigation into a step-wise inertial navigation and dead reckoning. 
This architecture does not completely solve the computational cost issue, but makes it manageable for up to a platoon-sized group of agents. For larger groups, some cellular structure is needed~\cite{Hawkinson2012,Martinelli2007}. However, the architecture is largely independent of how the global state estimation is implemented and a distributed implementation is conceivable.

The idea of dividing the filtering is not completely new. A similar division is presented in an application specific context in~\cite{Krach2008a} and used to fuse data from foot-mounted inertial sensors with maps, or to build the maps themselves, in~\cite{Angermann2012,Roberston2010,Pinchin2012}. However, the described division is heuristically motivated
and the statistical relation between the different parts is not clear. Also, no physical processing decentralization is exploited to give reduced communication requirements.
%



\subsection{Robust and low computational cost sensor fusion}\label{subsec:fusion}
The sensor fusion firstly poses the problem of how to model the relation between the tracked inertial sensors and the range measurements. Secondly, it poses the problem of how to condition the state statistic estimates on provided information while retaining reasonable computational costs. 


The easiest solution to the non-collocated sensors of individual agents is to make the assumption that they are collocated (or have a fixed relation)~\cite{Stromback2010,Rantakokko2012a,Godha2008,Bird2011}. 
While simple, this method can clearly introduce modeling errors resulting in suboptimal performance and questionable integrity.
Instead, explicitly modeling the relative motion of the feet has been suggested in~\cite{Bancroft2008}. However, making an accurate and general model of the human motion is difficult, to say the least. 
As an alternative, multiple publications suggest explicitly measuring the  relation between the sensors~\cite{Laverne2011,Brand2003,Hung2013,Saarinen2004}. The added information can improve the localization performance but unfortunately introduces the need for additional hardware and measurement models. 
Also, it works best for situations with line-of-sight between measurement points, and therefore, it is probably only a viable solution for foot-to-foot ranging on clear, not too rough, and vegetation/obstacle free ground~\cite{Zhang2007}. 
%
Instead of modeling or measuring the relation between navigation points of an individual agent, the constraint that the spatial separation between them has an upper limit may be used. This side information obviously has an almost perfect integrity, and results in~\cite{Zachariah2012} indicate that the performance loss in comparison to ranging is transitory. For inertial navigation, it has been demonstrated that a range constraint can be used to fuse the information from two foot-mounted systems, while only propagating the mean and the covariance~\cite{Skog2012,Prateek2013}. Unfortunately, the suggested methods depend on numerical solvers and only apply the constraint on the mean, giving questionable statistical properties. Therefore, in Section~\ref{sec:sensor_fusion}, based on the work in~\cite{Zachariah2012}, we suggest a simpler and numerically more attractive solution to using range constraints to perform the sensor fusion, based on marginalization and sampling.


The naive solution to the sensor fusion of the foot-mounted inertial navigation and the inter-agent ranging is simply using traditional Kalman filter measurement updates for the ranging~\cite{Stromback2010}. However, the radio ranging errors are often far from Gaussian, often with heavy tails and non-stationary and spatially correlated errors~\cite{Lee2011,Abrudan2011,Hashemi1993,Pahlavan2006,Yoon2011,Olson2006}. This can cause unexpected behavior of many localization algorithms, and therefore, statistically more robust methods are desirable~\cite{Whitehouse2006,Yoon2011,Olson2006}. 
See~\cite{Zoubir2012} and references therein for a general treatment of the statistical robustness concept.
The heavy tails and spatially correlated errors could potentially be solved by a flat likelihood function as suggested in~\cite{Lee2011,Zampella2012}. However, while giving a high integrity, this also ignores a substantial amount of information and requires multi-hypothesis filtering (a particle filter) with unacceptable high computational cost. 
Using a more informative likelihood function is not hard to imagine. Unfortunately, only a small set of likelihood functions can easily be used without resorting to multi-hypothesis filtering methods. Some low cost fusion techniques for special classes of heavy-tailed distributions and H$_\infty$ criteria have been suggested in the literature~\cite{Sornette2001,Vila2011,Gordon2003,Wang2008,Allen2010}. However, ideally we would like more flexibility to handle measurement errors and non-collocated sensors. Therefore, in Section~\ref{sec:sensor_fusion}, we propose a marginalization and sample based measurement update for the inter-agent ranging, providing the necessary flexibility to handle an arbitrary likelihood function. A suitable likelihood function is proposed, taking lack of collocation, statistical robustness, and correlated errors into account, and shown to providing a robust and low computational cost sensor fusion.



\section{Decentralized estimation architecture}\label{sec:sys_arch}
To get around the problems of the centralized architecture, the state estimation needs somehow to be partially decentralized. However, as previously argued, some global state estimation is necessary. Consequently, the challenge is to do the decentralization in a way that does not lead to unacceptable loss in information coupling, leading to poor performance and integrity, while still solving the issues with computational cost, communication bandwidth, and robustness to varying connectivity. In the following subsections it is shown how this can be achieved by dividing the filtering associated with foot-mounted inertial sensors into a step-wise inertial navigation and step-wise dead reckoning, as illustrated in Fig.~\ref{fig:tracking_ill}. 
Pseudo-code for the related processing is found in Alg.~\ref{Alg:recursive_segm} and the complete system architecture is illustrated in Fig.~\ref{fig:sys_arch}.

\subsection{Zero-velocity-update-aided inertial navigation}\label{S:FMINS}
To track the position of an agent equipped with foot-mounted inertial sensors, the sensors are used to implement an inertial navigation system aided by so called zero-velocity updates (ZUPTs).
%
%
The inertial navigation essentially consists of the inertial sensors combined with mechanization equations. In the simplest form, the mechanization equations are
\begin{equation}\label{eq:mechanization}
\begin{bmatrix}
\mathbf{p}_k \\
\mathbf{v}_k \\
\mathbf{q}_k
\end{bmatrix}
=
\begin{bmatrix}
\mathbf{p}_{k-1} + \mathbf{v}_{k-1}dt \\
\mathbf{v}_{k-1} + (\mathbf{q}_{k-1}\mathbf{f}_k\mathbf{q}^{\star}_{k-1}-\mathbf{g})dt\\
\boldsymbol{\Omega}(\boldsymbol{\omega}_kdt)\mathbf{q}_{k-1}
\end{bmatrix}
\end{equation}
where $k$ is a time index, $dt$ is the time difference between measurement instances, $\mathbf{p}_k$ is the position, $\mathbf{v}_k$ is the velocity, $\mathbf{f}_k$ is the specific force, $\mathbf{g}=[0,0,g]$ is the gravity, and $\boldsymbol{\omega}_k$ is the angular rate (all in $\mathbb{R}^3$). Further, $\mathbf{q}_k$ is the quaternion describing the orientation of the system, 
the triple product $\mathbf{q}_{k-1}\mathbf{f}_k\mathbf{q}^{\star}_{k-1}$ denotes the rotation of $\mathbf{f}_k$ by $\mathbf{q}_{k}$, and $\boldsymbol{\Omega}(\cdot)$ is the quaternion update matrix. For a detailed treatment of inertial navigation see~\cite{Britting1971,Jekeli2001}. For analytical convenience we will interchangeably represent the orientation $\mathbf{q}_k$ with the equivalent Euler angles (roll,pitch,yaw) ${\boldsymbol{\theta}}_{k}=[\phi_k,\theta_k,\psi_k]$. Note that $[\,\cdot\,,\dots]$ is used to denote a column vector.

The mechanization equations \eqref{eq:mechanization} together with measurements of the specific force $\tilde{\mathbf{f}}_k$ and the angular rates $\tilde{\boldsymbol{\omega}}_k$, provided by the inertial sensors, are used to propagate position $\hat{\mathbf{p}}_k$, velocity $\hat{\mathbf{v}}_k$, and orientation $\hat{\mathbf{q}}_k$ state estimates. Unfortunately, due to its integrative nature, small measurement errors in $\tilde{\mathbf{f}}_k$ and $\tilde{\boldsymbol{\omega}}_k$ accumulate, giving rapidly growing estimation errors. Fortunately, these errors can be modeled and estimated with ZUPTs. A first-order error model of \eqref{eq:mechanization} is given by
\begin{equation}\label{eq:deviation_model}
\begin{bmatrix}
\delta {\mathbf{p}}_{k}\\
\delta {\mathbf{v}}_{k}\\
\delta {\boldsymbol{\theta}}_{k}
\end{bmatrix}
=
\begin{bmatrix}
\mathbf{I} & \mathbf{I}dt & \mathbf{0}\\
\mathbf{0} & \mathbf{I} & [\mathbf{q}_{k-1}\mathbf{f}_k\mathbf{q}^{\star}_{k-1}]_\times dt \\
\mathbf{0} & \mathbf{0} & \mathbf{I}
\end{bmatrix}
\begin{bmatrix}
\delta {\mathbf{p}}_{k-1}\\
\delta {\mathbf{v}}_{k-1}\\
\delta {\boldsymbol{\theta}}_{k-1}
\end{bmatrix}
\end{equation}
where $\delta(\cdot)_k$ are the error states, $\mathbf{I}$ and $\mathbf{0}$ are $3\times 3$ identity and zero matrices, respectively, and $[\cdot]_\times$ is the cross-product matrix. 
As argued in~\cite{Nilsson2012a}, one should be cautious about estimating systematic sensor errors in the current setup. 
Indeed, remarkable dead-reckoning performance has been demonstrated, exploiting dual foot-mounted sensors without any sensor error state estimation~\cite{Kelly2011}. Therefore, in contrast to many publications, no additional sensor bias states are used.

Together with statistical models for the errors in $\tilde{\mathbf{f}}_k$ and $\tilde{\boldsymbol{\omega}}_k$, \eqref{eq:deviation_model} is used to propagate statistics of the error states. To estimate the error states, stationary time instances are detected based on the condition $Z(\{\tilde{\mathbf{f}}_\kappa,\tilde{\boldsymbol{\omega}}_\kappa\}_{W_k})<\gamma_\text{\textsc{z}}$, where $Z(\cdot)$ is some zero-velocity test statistic, $\{\tilde{\mathbf{f}}_\kappa,\tilde{\boldsymbol{\omega}}_\kappa\}_{W_k}$ is the inertial measurements over some time window $W_k$, and $\gamma_\text{\textsc{z}}$ is a zero-velocity detection threshold. See~\cite{Skog2010,Skog2010a} for further details. The implied zero-velocities are used as pseudo-measurements
\begin{equation}\label{eq:zupts_meas}
\tilde{\mathbf{y}}_k = \hat{\mathbf{v}}_k\:\forall k:Z(\{\tilde{\mathbf{f}}_\kappa,\tilde{\boldsymbol{\omega}}_\kappa\}_{W_k})<\gamma_\text{\textsc{z}}
\end{equation}
which are modeled in terms of the error states as
\begin{equation}\label{eq:zupts}
\tilde{\mathbf{y}}_k=\mathbf{H}
\begin{bmatrix}
\delta {\mathbf{p}}_{k}\\
\delta {\mathbf{v}}_{k}\\
\delta {\boldsymbol{\theta}}_{k}
\end{bmatrix}
 + \mathbf{n}_k
\end{equation}
where $\mathbf{H}=[\mathbf{0}\:\:\mathbf{I}\:\:\mathbf{0}]$ is the measurement matrix and $\mathbf{n}_k$ is a measurement noise, i.e. $\tilde{\mathbf{y}}_k=\delta\mathbf{v}_k + \mathbf{n}_k$. A similar detector is also used to lock the system when completely stationary. See~\cite{Nilsson2013b} for further details. Given the error model \eqref{eq:deviation_model} and the measurements model \eqref{eq:zupts}, the measurements \eqref{eq:zupts_meas} can be used to estimate the error states with a Kalman type of filter. See~\cite{Foxlin2005,Skog2010a,Bebek2010,Jimenez2010} for further details and variations. See~\cite{Farrell2008} for a general treatment of aided navigation. Since there is no reason to propagate errors, as soon as there are any non-zero estimates $\delta\hat{\mathbf{p}}_k$, $\delta\hat{\mathbf{v}}_k$, or $\delta\hat{\boldsymbol{\theta}}_k$, they are fed back correcting the navigational states,
\begin{equation}\label{eq:feedback}
\begin{bmatrix}
\hat{\mathbf{p}}_k\\
\hat{\mathbf{v}}_k
\end{bmatrix}
:=
\begin{bmatrix}
\hat{\mathbf{p}}_k\\
\hat{\mathbf{v}}_k
\end{bmatrix}
+
\begin{bmatrix}
\delta\hat{\mathbf{p}}_k\\
\delta\hat{\mathbf{v}}_k
\end{bmatrix}
\quad\text{and}\quad
\hat{\mathbf{q}}_k:=\boldsymbol{\Omega}(\delta\hat{\boldsymbol{\theta}}_k)\hat{\mathbf{q}}_k
\end{equation}
and consequently the error state estimates are set to zero, i.e. $\delta\hat{\mathbf{p}}_k:=\mathbf{0}_{3\times 1}$, $\delta\hat{\mathbf{v}}_k:=\mathbf{0}_{3\times 1}$, and $\delta\hat{\boldsymbol{\theta}}_k:=\mathbf{0}_{3\times 1}$, where $:=$ indicates assignment.

Unfortunately, all (error) states are not observable based on the ZUPTs. During periods of consecutive ZUPTs, 
the system \eqref{eq:deviation_model} becomes essentially linear and time invariant. 
Zero-velocity for consecutive time instances means no acceleration and ideally $\mathbf{f}_k=\mathbf{q}^{\star}_k\mathbf{g}\mathbf{q}_k$. This gives the system and observability matrices
\begin{equation*}
\mathbf{F} =
\begin{bmatrix}
\mathbf{I} & \mathbf{I}dt & \mathbf{0} \\
\mathbf{0} & \mathbf{I} & [\mathbf{g}]_\times dt \\
\mathbf{0} & \mathbf{0} & \mathbf{I}
\end{bmatrix}
\:\:\text{and}\:\:
\begin{bmatrix}
\mathbf{H} \\
\mathbf{H}\mathbf{F} \\
\mathbf{H}\mathbf{F}^2
\end{bmatrix}
=
\begin{bmatrix}
\mathbf{0} & \mathbf{I} & \mathbf{0} \\
\mathbf{0} & \mathbf{I} & [\mathbf{g}]_\times dt \\
\mathbf{0} & \mathbf{I} & 2[\mathbf{g}]_\times dt
\end{bmatrix}\!.
\end{equation*}
Obviously, the position (error) is not observable, while the velocity is.
Since
\begin{equation*}
[\mathbf{g}]_\times =
\begin{bmatrix}
0 & g & 0 \\
-g & 0 & 0 \\
0 & 0 & 0
\end{bmatrix}
\end{equation*}
the roll and pitch are observable while the heading (yaw) of the system is not. 
Ignoring the process noise, this implies that the covariances of the observable states decay as one over 
the number of consecutive ZUPTs. Note that 
there is no difference between the covariances of the error states and the states themselves.
Consequently, during stand-still, after a reasonable number of ZUPTs, the state estimate covariance becomes
\begin{equation}\label{eq:zupt_cov}
\text{cov}\left((\hat{\mathbf{p}}_{k},\hat{\mathbf{v}}_{k},\hat{\boldsymbol{\theta}}_{k})\right)\!\approx\!\!
\begin{bmatrix}
\mathbf{P}_{\mathbf{p}_k} & \mathbf{0}_{3\times 5} & \mathbf{P}_{\mathbf{p}_k,\psi_k} \\
\mathbf{0}_{5\times 3} & \mathbf{0}_{5\times 5} & \mathbf{0}_{5\times 1} \\
\mathbf{P}^\top_{\mathbf{p}_k,\psi_k} & \mathbf{0}_{1\times 5} & \mathbf{P}_{\psi_k,\psi_k}
\end{bmatrix}
\end{equation}
where $\mathbf{P}_{x,y}\!=\!\text{cov}(x,\!y)$, $\mathbf{P}_x\!=\!\text{cov}(x)\!=\!\text{cov}(x,\!x)$, $(\cdot)^\top$ denotes the transpose, and $\mathbf{0}_{n\times m}$ denotes a zero matrix of size $n\times m$.

\subsection{Step-wise dead reckoning}\label{subsec:swdr}
The covariance matrix \eqref{eq:zupt_cov} tells us that the errors of $\hat{\mathbf{p}}_k$ and $\hat{\psi}_k$ are uncorrelated with those of $\hat{\mathbf{v}}_k$ and $[\hat{\phi}_k,\hat{\theta}_k]$. Together with the Markovian assumption of the state space models and the translational and in-plan rotation invariance of \eqref{eq:mechanization}-\eqref{eq:zupts}, this means that future errors of $\hat{\mathbf{v}}_k$ and $[\hat{\phi}_k,\hat{\theta}_k]$ are uncorrelated with those of the current $\hat{\mathbf{p}}_k$ and $\hat{\psi}_k$. Consequently, future ZUPTs cannot be used to deduce information about the current position and heading errors. In turn, this means that, considering only the ZUPTs, it makes no difference if we reset the system and add the new relative position and heading to those before the reset. However, for other information sources, we must keep track of the global (total) error covariance of the position and heading estimates. 

Resetting the system means setting position $\hat{\mathbf{p}}_k$ and heading $\hat{\psi}_k$ and corresponding covariances to zero. Denote the position and heading estimates at a reset $\ell$ by $d\mathbf{p}_\ell$ and $d\psi_\ell$. 
These values can be used to drive the step-wise dead reckoning
%
%
%
\begin{equation}\label{eq:simple_ss}
\begin{bmatrix}
\mathbf{x}_\ell \\
\chi_\ell
\end{bmatrix}
=
\begin{bmatrix}
\mathbf{x}_{\ell-1} \\
\chi_{\ell-1}
\end{bmatrix}
+
\begin{bmatrix}
\mathbf{R}_{\ell-1}d\mathbf{p}_\ell \\
d\psi_\ell
\end{bmatrix}
+
\mathbf{w}_\ell
\end{equation}
where $\mathbf{x}_\ell$ and $\chi_\ell$ are the global position in 3 dimensions and heading in the horizontal plan of the inertial navigation system relative to the navigation frame,
\begin{equation*}
\mathbf{R}_\ell=
\begin{bmatrix}
\cos(\chi_\ell) & -\sin(\chi_\ell) & 0 \\
\sin(\chi_\ell) & \cos(\chi_\ell) & 0 \\
0 & 0 & 1
\end{bmatrix}
\end{equation*}
is the rotation matrix from the local coordinate frame of the last reset to the navigation frame, and $\mathbf{w}_\ell$ is a (by assumption) white noise with covariance,
\begin{align}
\text{cov}(\mathbf{w}_\ell) =&\, \text{cov}\big(\,[\mathbf{R}_{\ell-1}d\mathbf{p}_\ell,d\psi_\ell]\,\big)\nonumber\\
                   =&
\begin{bmatrix}
\mathbf{R}_{\ell-1}\mathbf{P}_{\mathbf{p}_\ell}\mathbf{R}^{\top}_{\ell-1} & \mathbf{R}_{\ell-1}\mathbf{P}_{\mathbf{p}_\ell,\psi_\ell} \\
\mathbf{P}^\top_{\mathbf{p}_\ell,\psi_\ell}\mathbf{R}^{\top}_{\ell-1} & P_{\psi_\ell,\psi_\ell}
\end{bmatrix}\!.\label{eq:cov_dr}
\end{align}
The noise $\mathbf{w}_\ell$ in \eqref{eq:simple_ss} represents the accumulated uncertainty in position and heading since the last reset, i.e. the essentially non-zero elements in \eqref{eq:zupt_cov} transformed to the navigation frame. 
The dead reckoning \eqref{eq:simple_ss} can trivially be used to estimate $\mathbf{x}_\ell$ and $\chi_\ell$ and their error covariances from $d\mathbf{p}_\ell$ and $d\psi_\ell$ and related covariances. The relation between the step-wise inertial navigation and dead reckoning is illustrated in Fig.~\ref{fig:tracking_ill}.


%

\begin{figure}[t]
\centering
{\resizebox{\linewidth}{!}{\includegraphics{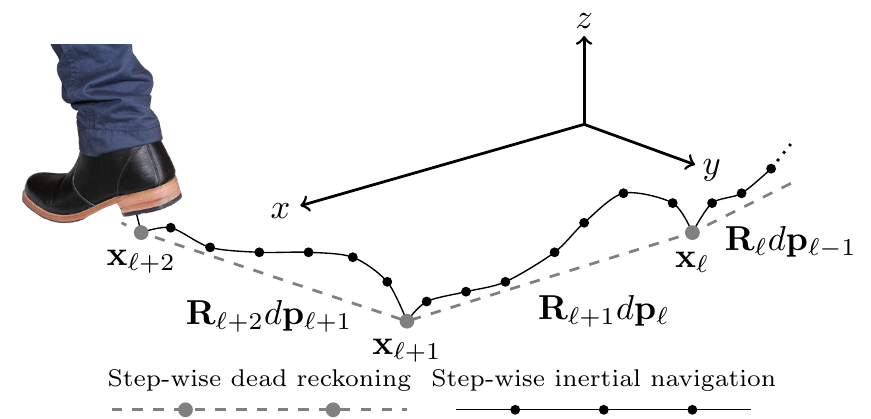}}}
\caption{Illustration of the step-wise inertial navigation and the step-wise dead reckoning. The displacement and heading change over a step given by the inertial navigation is used to perform the step-wise dead-reckoning.}\label{fig:tracking_ill}
\end{figure}

To get $[d\mathbf{p}_\ell,d\psi_\ell]$ from the inertial navigation, reset instances need to be determined, i.e. the decoupled situation \eqref{eq:zupt_cov} needs to be detected.
However, detecting it is not enough. If it holds for one time instance $k$, it is likely to hold for the next time instance. Resetting at nearby time instances is not desirable. 
Instead we want to reset once at every step or at some regular intervals if the system is stationary for a longer period of time. The latter requirement is necessary to distinguish between extended stationary periods and extended dynamic periods. 
Further, to allow for real-time processing, the detection needs to be done in a recursive manner. 
The longer the stationary period, the smaller the cross-coupling terms in \eqref{eq:zupt_cov}. This means that the system should be reset as late as possible in a stationary period. 
However, if the stationary period is too short, we may not want to reset at all, since then the cross terms in \eqref{eq:zupt_cov} may not have converged. 
In summary, \emph{necessary conditions} for a reset are low enough cross-coupling and minimum elapsed time since the last reset. If this holds, there is a pending reset.
In principle, the cross-coupling terms in \eqref{eq:zupt_cov} should be used to determine the first requirement. However, in practice, all elements fall off together and a threshold $\gamma_p$ on e.g. the first velocity component can be used.
To assess the second requirement, a counter $c_{p}$ which is incremented at each time instance is needed, giving the pending reset condition
\begin{equation}\label{eq:pending_reset}
(\mathbf{P}_{v_{x_k}}<\gamma_p) \wedge (c_{p}>c_{\min})
\end{equation}
where $c_{\min}$ is the minimum number of samples between resets.
A pending reset is to be performed if the stationary period comes to an end or a maximum time with a pending reset has elapsed. To assess the latter condition, a counter $c_{d}$ is needed which is incremented if \eqref{eq:pending_reset} holds. Then a reset is performed if
\begin{equation}\label{eq:rec_seg}
\big(Z(\{\tilde{\mathbf{f}}_\kappa,\tilde{\boldsymbol{\omega}}_\kappa\}_{W_k})\geq \gamma_\text{\textsc{z}}\big) \vee (c_{d}>c_{\max})
\end{equation}
where $c_{\max}$ is the maximum number of samples of a pending reset. Together \eqref{eq:pending_reset} and \eqref{eq:rec_seg} make up the \emph{sufficient conditions} for a reset. When the reset is performed, the counters are reset, $c_{p}:= 0$ and $c_{d}:= 0$. This gives a recursive step segmentation. Pseudo-code for the inertial navigation with recursive step segmentation (i.e. step-wise inertial navigation) and the step-wise dead reckoning is found in Alg.~\ref{Alg:recursive_segm}. 

\begin{algorithm}[t]
\caption{\footnotesize Pseudo-code of the combined step-wise inertial navigation and step-wise dead reckoning. The ZUPT-aided inertial navigation and the step-wise dead reckoning refers to the effect of \eqref{eq:mechanization}-\eqref{eq:feedback} and \eqref{eq:simple_ss}, respectively, combined with Kalman type of filtering. For notational compactness, below $\mathbf{P}_k=\mathbf{P}_{[\mathbf{p}_k,\mathbf{v}_k,\mathbf{q}_k]}$ and $\mathbf{P}_\ell=\mathbf{P}_{[\mathbf{x}_\ell,\!\chi_\ell]}$.}\label{Alg:recursive_segm}
\begin{algorithmic}[1]
\STATE $k:=\ell:=c_p:=c_d:=0$
\STATE $\mathbf{p}_k:=\mathbf{v}_k:=\mathbf{0}_{3\times1}$
\STATE $\mathbf{q}_k:=\{\text{\small Coarse self-initialization}\}$ (See e.g.~\cite{Farrell2008})
\STATE $\mathbf{P}_k:=\{\text{\small Initial velocity, roll and pitch uncertainty}\}$
\STATE $(\mathbf{x}_\ell,\chi_\ell):=\{\text{\small Initial position and heading}\}$\label{alg:swdr_init}
\STATE $\mathbf{P}_{(\mathbf{x}_\ell,\chi_\ell)}:=\{\text{\small Initial position and heading uncertainty}\}$
\LOOP
\STATE $k:=k+1$
\STATE {\small ZUPT-aided inertial navigation}\\
$([\mathbf{p}_k,\mathbf{v}_k,\mathbf{q}_k],\mathbf{P}_k)\!\leftarrow\!([\mathbf{p}_{k-1},\mathbf{v}_{k-1},\mathbf{q}_{k-1}],\!\mathbf{P}_{k-1},\!\tilde{\mathbf{f}}_k,\!\tilde{\boldsymbol{\omega}}_k\!)$
\STATE $c_{p}:=c_{p}+1$
\IF {$(P_{v_k}<\gamma_p) \wedge (c_{p}>c_{\min})$}
        \STATE $c_{d}:=c_{d-1}+1$
        \IF {$\big(Z(\{\tilde{\mathbf{f}}_\kappa,\tilde{\boldsymbol{\omega}}_\kappa\}_{W_k})\geq \gamma_\text{\textsc{z}}\big) \vee (c_{d}>c_{\max})$}
            \STATE $\ell:=\ell+1$
            \STATE $d\mathbf{p}_\ell:=\hat{\mathbf{p}}_k,\:d\psi_\ell:=\hat{\phi}_k,\:\mathbf{P}_{w_\ell}=\dots$ (see~\eqref{eq:cov_dr})
            \STATE $\mathbf{p}_k:=\mathbf{v}_k:=\mathbf{0}_{3\times1},\:\psi_k:=0$
            \STATE $\mathbf{P}_k:=\mathbf{0}_{9\times9}$
            \STATE $c_{p}:= 0,\:c_{d}:= 0$
            \STATE {\small Step-wise dead reckoning}\label{alg:swdr}\\ $([\mathbf{x}_\ell,\chi_\ell],\!\mathbf{P}_\ell)\!\leftarrow\!([\mathbf{x}_{\ell-1},\!\chi_{\ell-1}],\!\mathbf{P}_{\ell-1},\!d\mathbf{p}_\ell,\!d\psi_\ell,\!\mathbf{P}_{w_\ell})$
        \ENDIF
\ENDIF
\ENDLOOP
\end{algorithmic}
\end{algorithm}

Not to lose performance in comparison with a sensor fusion approach based on centralized estimation, the step-wise inertial navigation combined with the step-wise dead reckoning needs to reproduce the same state statistics (mean and covariance) as those of the indefinite (no resets) ZUPT-aided inertial navigation. If the models \eqref{eq:mechanization}, \eqref{eq:deviation_model}, and \eqref{eq:simple_ss} had been linear with Gaussian noise and the cross-coupling terms of \eqref{eq:zupt_cov} were perfectly zero, then the divided filtering would reproduce the full filter behavior perfectly. Unfortunately, they are not.
However, as shown in the example trajectory in Fig.~\ref{fig:consistency}, in practice the differences are marginal and the mean and covariance estimates of the position and heading can be reproduced by only $[d\mathbf{p}_\ell,d\psi_\ell]$ and the corresponding covariances. Due to linearization and modeling errors of the ZUPTs, the step-wise dead reckoning can even be expected to improve performance since it will eliminate these effects to single steps~\cite{Nilsson2012a,Bageshwar2009}. Indeed, resetting appropriate covariance elements (which has similar effects as of performing the step-wise dead reckoning) has empirically been found to improve performance~\cite{Rubia2013}.

\begin{figure}[t]
\centering
{\resizebox{0.95\linewidth}{!}{\includegraphics{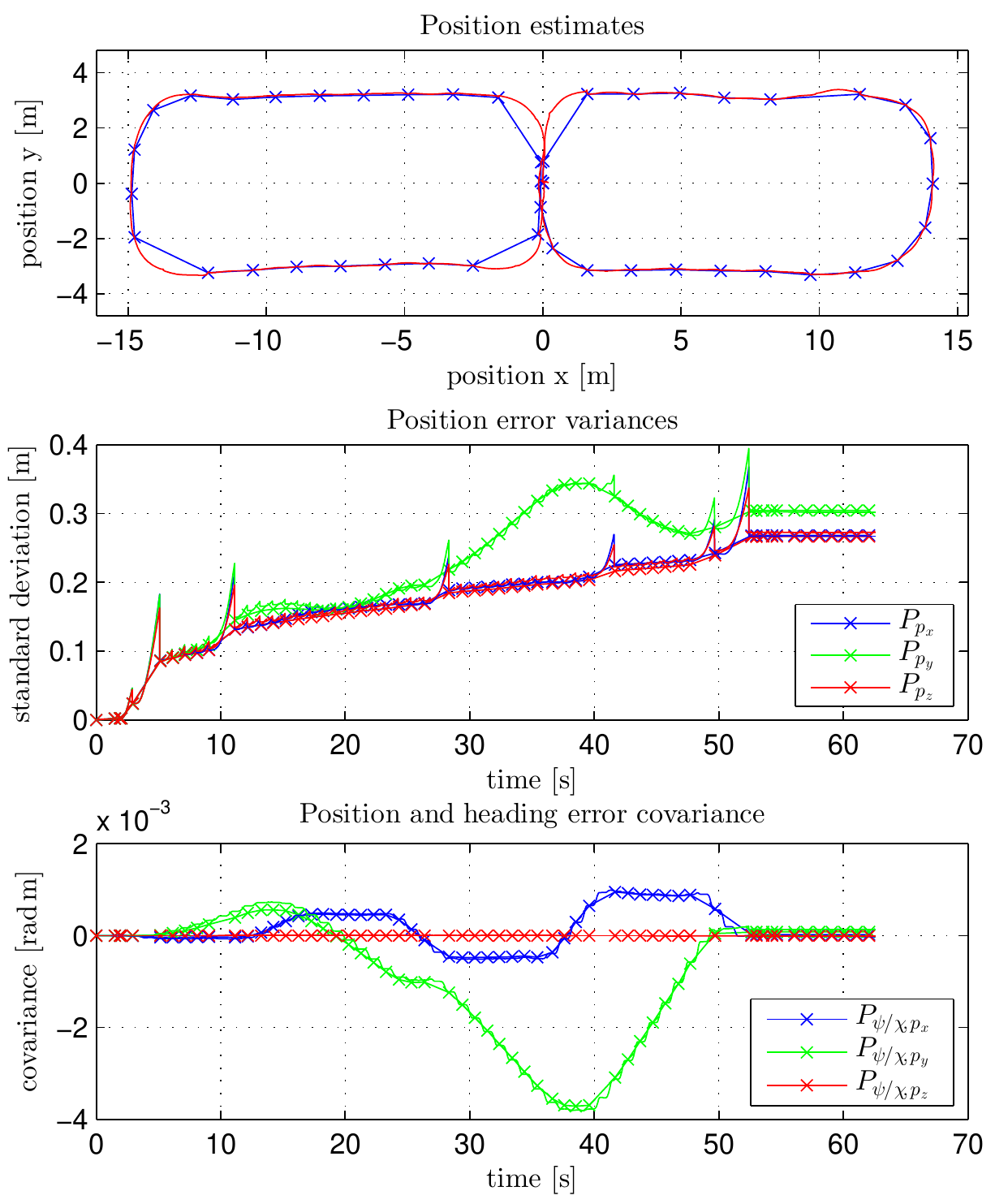}}}
\caption{The plots show the trajectory (upper), the position error covariances (middle), and the covariance between the position and heading errors (lower) as estimated by an extended Kalman filter based indefinite ZUPT-aided inertial navigation (solid lines) and a step-wise inertial navigation and dead reckoning (crossed lines). The agreement between the systems are far below the accuracy and integrity of the former system.}\label{fig:consistency}
\end{figure}

\subsection{Physical decentralization of state estimation}\label{sec:phys_arc}
The step-wise inertial navigation and dead reckoning as described in Alg.~\ref{Alg:recursive_segm} can be used to implement a decentralized architecture and state estimation. The ranging, as well as most additional information, is only dependent on position and not on the full state vector $[\mathbf{p}_k,\mathbf{v}_k,\boldsymbol{\theta}_k]$. Further, as argued in the previous subsection, the errors of $\hat{\mathbf{v}}_k$ and $[\hat{\phi}_k,\hat{\theta}_k]$ are weakly correlated with those of $\hat{\mathbf{p}}_k$ and $\hat{\psi}_k$. Therefore, only the states $[\mathbf{x}_\ell,\chi_\ell]$ (for all feet) have to be estimated jointly and only line \ref{alg:swdr} need to be executed centrally. 
The step-wise inertial navigation, i.e. Alg.~\ref{Alg:recursive_segm} apart from line \ref{alg:swdr}, can be implemented locally in the foot-mounted units, and thereby, only $[d\mathbf{p}_\ell,d\psi_\ell]$ and related covariances need to be transmitted from the feet. This way, the required communication will be significantly lower compared to the case in which all inertial data would have to be transmitted. Also, since the computational cost of propagating \eqref{eq:simple_ss} is marginal, this can be done both locally on the processing device of each agent and in a global state estimation. This way, if an agent loses contact with whomever who performs the global state estimation, it can still perform the dead reckoning, and thereby, keep an estimate of where it is. Since the amount of data in the displacement and heading changes is small, if contact is reestablished, all data can easily be transferred and its states in the global state estimation updated. The other way around, if corrections to the estimates of $[\mathbf{x}_\ell,\chi_\ell]$ are made in the central state estimation, these corrections can be transferred down to the agent. Since the recursion in \eqref{eq:simple_ss} is pure dead reckoning (no statistical conditioning), these corrections can directly be used to correct the local estimates of $[\mathbf{x}_\ell,\chi_\ell]$. This way, the local and the global estimates can be kept consistent.
%
%

The straight-forward way of implementing the global state estimation is by one (or multiple) central fusion center to which all dead reckoning data are transmitted (potentially by broadcasting). The fusion center may be carried by an agents, or reside in a vehicle or something similar. Range measurements relative to other agents only have a meaning if the position estimate and its statistics are known. Therefore, all ranging information is transferred to the central fusion center. This processing architecture with its three layers of estimation (foot, processing device of agent, and common fusion center) is illustrated in Fig.~\ref{fig:sys_arch}. However, the division in step-wise inertial navigation and dead reckoning is independent of the structure with a fusion center and some decentralized global state estimation could potentially be used.
%

\begin{figure}[t]
\centering
{\resizebox{\linewidth}{!}{\includegraphics{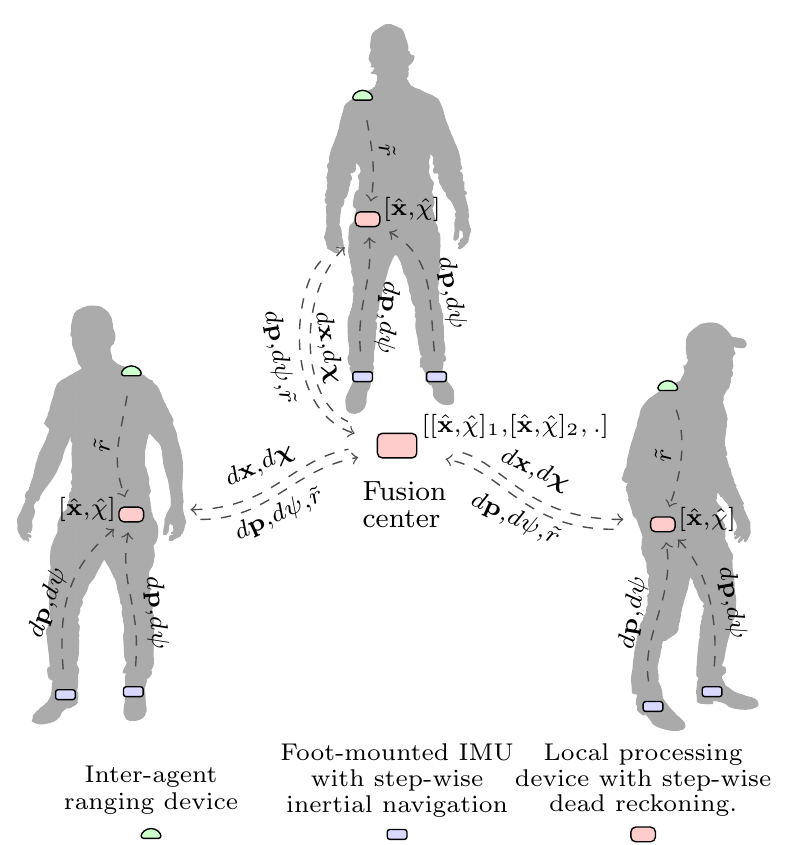}}}
\caption{Illustration of the decentralized system architecture. Step-wise inertial navigation is done locally in the foot-mounted units. Displacement and heading changes are transferred to a local processing device, where step-wise dead reckoning is performed, and relayed together with ranging data to a central fusion center. The fusion center may be carried by an agents, reside in a vehicle or something similar, or be distributed among agents.}\label{fig:sys_arch}
\end{figure}

\subsection{Computational cost and required communication}\label{sec:com}
The step-wise dead reckoning is primarily motivated and justified by the reduction in computational cost and required communication bandwidth. 
With a completely centralized sensor fusion $[\tilde{\mathbf{f}}_k,\tilde{\boldsymbol{\omega}}_k]$, 6 measurement values in total, need to be transferred to the central fusion center at a sampling rate $f_\text{IMU}$ in the order of hundreds of [Hz] with each measurement value typically consisting of some 12-16 bits. With the step-wise dead reckoning, $[d\mathbf{p}_\ell,d\psi_\ell]$, $\mathbf{P}_{\mathbf{p}_\ell}$, $\mathbf{P}_{\mathbf{p}_\ell,\psi_\ell}$, and $P_{\psi_\ell,\psi_\ell}$, in total 14 values, need to be transferred to the central fusion center at a rate of $f_{\text{sw}}\approx 1$~[Hz] (normal gait frequency~\cite{Oberg1993}). In practice, the 14 values can be reduced to 8 values since cross-covariances may be ignored and the numerical ranges are such that they can reasonably be fitted in some 12-16 bits each. The other way around, some 4 correction values need to be transferred back to the agent.
Together this gives the ratio of the required communication of $(6\cdot f_\text{IMU})/(12\cdot f_\text{sw})\approx 10^{2}$, a two orders magnitude reduction. 


In turn, the computational cost scales linearly with the update rates $f_\text{IMU}$ and $f_\text{sw}$. 
In addition, the computational cost has a cubic scaling (for covariance based filters) with the state dimensionality.
Therefore, the reduction in the computational cost at the central fusion center is at the most $f_\text{IMU}/f_\text{sw}\cdot\left(\nicefrac{9}{4}\right)^{3}\approx 10^3$. However, at higher update rates, updates may be bundle together. 
Consequently, a somewhat lower reduction may be expected in practice giving a reduction of again around two orders of magnitude.


\section{Robust and low-cost sensor fusion}\label{sec:sensor_fusion}
The step-wise dead reckoning provides a low dimensional and low update rate interface to the foot-mounted inertial navigation. With this interface, the global state of the localization system (the system state as conceived by the global state estimation) becomes
\begin{equation*}
\mathbf{x}=[\mathbf{x}_\alpha,\chi_\alpha,\mathbf{x}_\beta,\chi_\beta,\mathbf{x}_\zeta,\chi_\zeta,\dots]
\end{equation*}
where $\mathbf{x}_j$ and $\chi_j$ are the positions and headings of the agents' feet with dropped time indices. Other auxiliary states may also be put in the state vector. 
Our desire is to fuse the provided dead-reckoning with that of the other foot and that of other agents via inter-agent ranging. This fusion is primarily challenging because: 1) The high dimensionality of the global system. 2) The non-collocated sensors of individual agents. 3) The potentially malign error characteristic of the ranging.
The high dimensionality is tackled by only propagating mean and covariance estimates and by marginalization of the state space. 
The lack of collocation is handled by imposing range constraints between sensors. Finally, the error characteristic of the ranging is handled by sampling based updates. In the following subsections, these approaches are described. Pseudo-code for the sensor fusion is found in Algs.~\ref{Alg:range_constraint}-\ref{Alg:range_ud} in the final subsection.

\subsection{Marginalization}
New information (e.g. range measurements) introduced in the systems is only dependent on a small subset of the states. 
%
Assume that the state vector can be decomposed as $\mathbf{z}=[\mathbf{z}_1,\mathbf{z}_2]$, such that some introduced information $\pi$ is only dependent on $\mathbf{z}_1$. Then with a Gaussian prior with mean and covariance
\begin{equation}\label{eq:z_mean_and_cov}
\hat{\mathbf{z}}=
\begin{bmatrix}
\hat{\mathbf{z}}_1 \\ \hat{\mathbf{z}}_2
\end{bmatrix}
\quad\text{and}\quad
\mathbf{P}_\mathbf{z}=
\begin{bmatrix}
\mathbf{P}_{z_1} & \mathbf{P}_{z_1z_2}\\
\mathbf{P}^\top_{z_1z_2} & \mathbf{P}_{z_2}
\end{bmatrix},
\end{equation}
the conditional (with respect to $\pi$) mean of $\mathbf{z}_2$ and the conditional covariance can be expressed as~\cite{Zachariah2012}
\begin{equation}\label{eq:marginalization}
\begin{split}
\hat{\mathbf{z}}_{2|\pi}\!=\,&\mathbf{V}+\mathbf{U}\,\hat{\mathbf{z}}_{1|\pi}\\
\mathbf{P}_{z_1|\pi}\!=\,&\mathbf{C}_{z_1|\pi}-\hat{\mathbf{z}}_{1|\pi}\hat{\mathbf{z}}^\top_{1|\pi}\\
\mathbf{P}_{z_2|\pi}\!=\,&\mathbf{P}_{z_2}\!\!-\!\mathbf{U}\mathbf{P}_{z_1z_2}\!\!\!+\!\!\mathbf{V}\mathbf{V}^\top\!\!\!+\!\mathbf{Z}\!+\!\mathbf{U}\mathbf{C}_{z_1|\pi}\mathbf{U}^\top\!\!\!-\!\hat{\mathbf{z}}_{2|\pi}\hat{\mathbf{z}}^\top_{2|\pi}\!\!\\
\mathbf{P}_{z_1z_2|\pi}\!=\!\!\!\!&\,\,\,\,\,\hat{\mathbf{z}}_{1|\pi}\mathbf{V}^\top+\mathbf{C}_{z_1|\pi}\mathbf{U}^\top-\hat{\mathbf{z}}_{1|\pi}\hat{\mathbf{z}}^\top_{2|\pi}
\end{split}
\end{equation}
where $\mathbf{U}=\mathbf{P}_{z_1z_2}^\top\mathbf{P}^{-1}_{z_1}$, $\mathbf{V}=\hat{\mathbf{z}}_{2}-\mathbf{U}\hat{\mathbf{z}}_1$, $\mathbf{Z}=\mathbf{U}\,\hat{\mathbf{z}}_{1|\pi}\mathbf{V}^\top+\mathbf{V}\,\hat{\mathbf{z}}^\top_{1|\pi}\mathbf{U}^\top$, and $\mathbf{C}_{z_1|\pi}$ is the conditional second order moment of $\mathbf{z}_1$. 
Note that this will hold for any information $\pi$ only dependent on $\mathbf{z}_1$, not just range constraints as studied in~\cite{Zachariah2012}. Consequently, the relations \eqref{eq:marginalization} give a desired marginalization. To impose the information $\pi$ on \eqref{eq:z_mean_and_cov}, only the first and second conditional moments, $\hat{\mathbf{z}}_{1|\pi}$ and $\mathbf{C}_{z_1|\pi}$, need to be calculated. If $\pi$ is linearly dependent on $\mathbf{z}_1$ and with Gaussian errors, this will be equivalent with a Kalman filter measurement update. This may trivially be used to introduce information about individual agents. However, as we will show in the following subsections, this can also be used to couple multiple navigation points of individual agents without any further measurement and to introduce non-Gaussian ranging between agents. 

\subsection{Fusing dead reckoning from dual foot-mounted units}\label{subsec:constraint}
The position of the feet $\mathbf{x}_a$ and $\mathbf{x}_b$ of an agent (in general two navigation points of an agent) has a bounded spatial separation. This can be used to fuse the related dead reckoning without any further measurements. In practice, the constraint will often have different extents in the vertical and the horizontal direction. This can be expressed as a range constraint
\begin{equation}\label{eq:range_constraint}
\|\mathbf{D}_\gamma(\mathbf{x}_a-\mathbf{x}_b)\|\leq\gamma_{xy}.
\end{equation}
where $\mathbf{D}_\gamma$ is a diagonal scaling matrix with $\gamma=[1,1,\gamma_{xy}/\gamma_z]$ on the diagonal, and $\gamma_{xy}$ and $\gamma_z$ are the constraints in the horizontal and vertical direction. Unfortunately, there is no standard way of imposing such a constraint in a Kalman like framework~\cite{Simon2010}. 
Also, the position states being in arbitrary locations in the global state vector, i.e. $\mathbf{x}=[\dots,\mathbf{x}_a,\dots,\mathbf{x}_b,\dots]$, means that the state vector is not on the form of $\mathbf{z}$. Further, since the constraint \eqref{eq:range_constraint} has unbounded support, the conditional means $\left[\hat{\mathbf{x}}_{a|\gamma},\hat{\mathbf{x}}_{b|\gamma}\right]$ and covariances $\text{cov}\left([\mathbf{x}_{a|\gamma},\mathbf{x}_{b|\gamma}]\right)$ cannot easily be evaluated. Moreover, since the states are updated asynchronously (steps occurring at different time instances), the state estimates $\hat{\mathbf{x}}_a$ and $\hat{\mathbf{x}}_b$ may not refer to the same time instance.
The latter problem can be handled by adjusting the constraint $\gamma_{xy}$ by the time difference of the states. In principle, this means that an upper limit on the speed by which an agent moves is imposed. The former problems can be solved with the state transformation
\begin{equation}\label{eq:trans_matrix_con}
\mathbf{z} = \mathbf{T}_{\gamma}\mathbf{x}\quad\text{where}\quad
\mathbf{T}_{\gamma} = \left(
\begin{bmatrix}
\mathbf{D}_\gamma & -\mathbf{D}_\gamma\\
\mathbf{D}_\gamma & \mathbf{D}_\gamma
\end{bmatrix}
\oplus\mathbf{I}_{m-6}\right)\boldsymbol{\Pi}
\end{equation}
where $\mathbf{I}_{m-6}$ is the identity matrix of size $m-6$, $m$ is the dimension of $\mathbf{x}$, $\oplus$ denotes the direct sum of matrices, $\boldsymbol{\Pi}$ is a permutation matrix fulfilling
\begin{equation*}\label{eq:perm_matrix}
\boldsymbol{\Pi}[\dots,\mathbf{x}_a,\dots,\mathbf{x}_b,\dots]=[\mathbf{x}_a,\mathbf{x}_b,\dots]
\end{equation*}
and $\mathbf{z}_1=\mathbf{D}_\gamma(\mathbf{x}_a-\mathbf{x}_b)$.
%
%
%
With the state transformation $\mathbf{T}_{\gamma}$, the mean and covariance of $\mathbf{z}$ become $\hat{\mathbf{z}}=\mathbf{T}_{\gamma}\hat{\mathbf{x}}$ and $\mathbf{P}_{z}=\mathbf{T}_{\gamma}\mathbf{P}_x\mathbf{T}_{\gamma}^\top$. Inversely, since $\mathbf{T}_{\gamma}$ is invertible, the conditional mean and covariance of $\mathbf{x}$ become $\hat{\mathbf{x}}_{|\gamma}=\mathbf{T}_{\gamma}^{-1}\hat{\mathbf{z}}_{|\gamma}$ and $\mathbf{P}_{x|\gamma}=\mathbf{T}_{\gamma}^{-1}\mathbf{P}_{z|\gamma}\mathbf{T}_{\gamma}^{-\top}$. Therefore, if $\hat{\mathbf{z}}_{1|\gamma}$ and $\mathbf{C}_{z_1|\gamma}$ are evaluated, \eqref{eq:marginalization} gives $\hat{\mathbf{z}}_{|\gamma}$ and $\mathbf{P}_{z|\gamma}$ and thereby also $\hat{\mathbf{x}}_{|\gamma}$ and $\mathbf{P}_{x|\gamma}$. Fortunately, with $\mathbf{z}_1=\mathbf{D}_\gamma(\mathbf{x}_a-\mathbf{x}_b)$, the constraint \eqref{eq:range_constraint} becomes
\begin{equation*}
\|\mathbf{z}_1\|\leq\gamma_{xy}.
\end{equation*}
%
In contrast to \eqref{eq:range_constraint}, this constraint has a bounded support. Therefore, as suggested in~\cite{Zachariah2012}, the conditional means can be approximated by sampling and projecting sigma points.
\begin{equation*}
\hat{\mathbf{z}}_{1|\gamma}\approx\medop\sum_{i}w^{i}\mathbf{z}_1^{(i)}\quad\text{and}\quad \mathbf{C}_{z_1|\gamma}\approx\medop\sum_{i}w^{i}\mathbf{z}_1^{(i)}(\mathbf{z}_1^{(i)})^\top
\end{equation*}
where $\approx$ denotes approximate equality and 
\begin{equation}\label{eq:projection_spherical}
\mathbf{z}_1^{(i)}=
\begin{cases}
\mathbf{s}^{(i)}, & \|\mathbf{s}^{(i)}\|\leq\gamma_{xy}\\
\tfrac{\gamma_{xy}}{\|\mathbf{s}^{(i)}\|}\mathbf{s}^{(i)}, & 1\\
\end{cases}.
\end{equation}
Here $\mathbf{s}^{(i)}$ and $w^{(i)}$ are sigma points and weights
\begin{equation*}
\{\mathbf{s}^{(i)},w^{(i)}\}=
\left\{\!\!\!
\begin{array}{lll}
\{\hat{\mathbf{z}}_1& \!\!\!\!\!\!\!\,,1\!-\!\nicefrac{3}{\eta}\}, & i=0\\
\{\hat{\mathbf{z}}_1+\eta^{1/2}\mathbf{l}_i& ,\nicefrac{1}{2\eta}\}, & i\in[1,\dots,3]\\
\{\hat{\mathbf{z}}_1-\eta^{1/2}\mathbf{l}_{i-3}\!\!& ,\nicefrac{1}{2\eta}\}, & i\in[4,\dots,6]\\
\end{array}\right.
\end{equation*}
where $\mathbf{l}_i$ is the $i$th column of the Cholesky decomposition $\mathbf{P}_{z_1}=\mathbf{L}\mathbf{L}^\top$ and the scalar $\eta$ reflects the portion of the prior to which the constraint is applied. 
See~\cite{Zachariah2012} for further details. The application of the constraint for the two-dimensional case is illustrated in Fig.~\ref{fig:ill_rcud}.

\begin{figure}[t]
\centering
{\resizebox{0.98\linewidth}{!}{\includegraphics{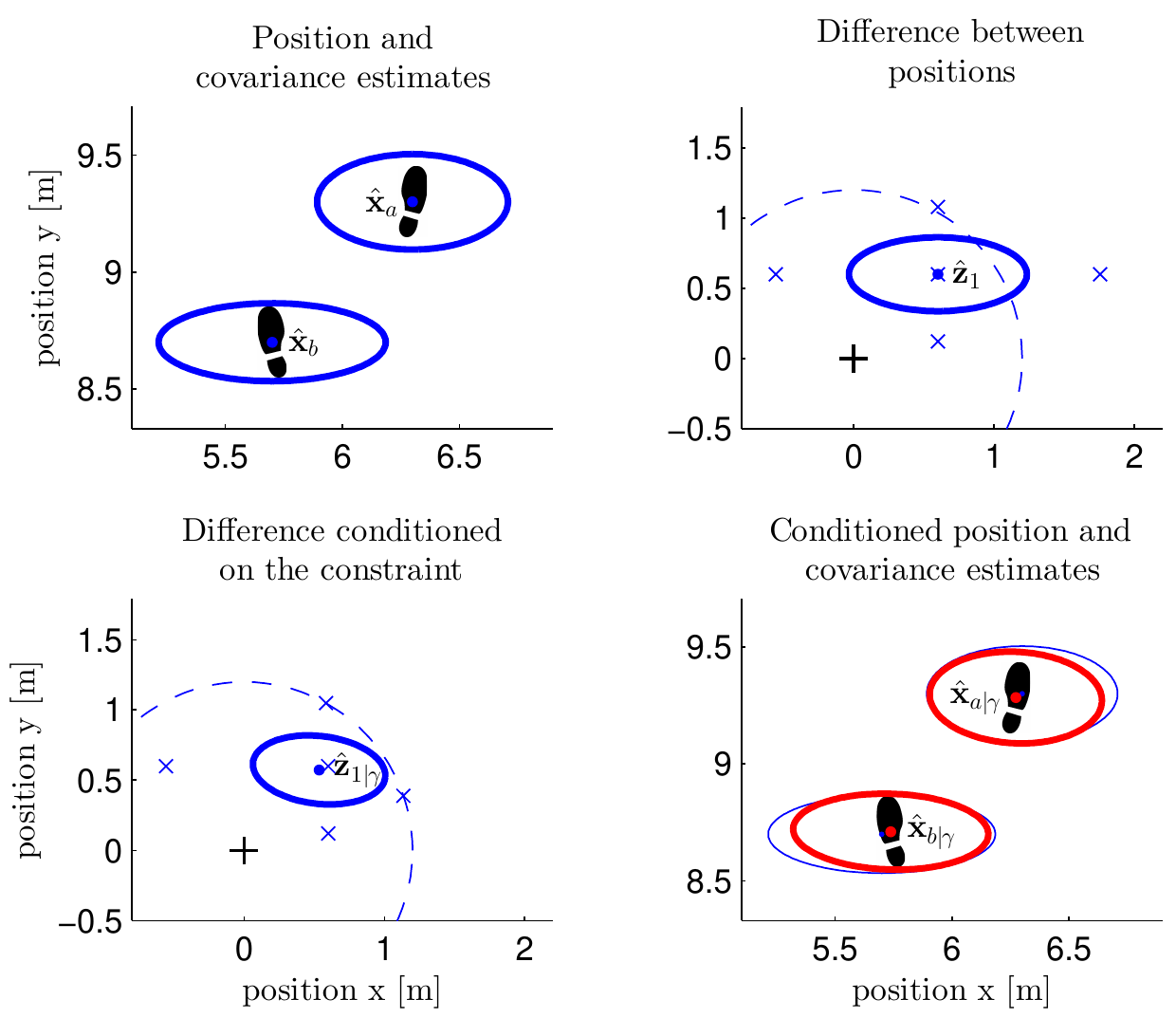}}}
\caption{Illustration of the separation constraint update for two feet in the horizontal plane. The plots show (upper-left) the prior position and covariance estimates, (upper-right) the transformed system with the sigma points (blue crosses) and the constraint indicated with the dashed circle, (lower-left) the projected sigma points and the conditional mean and covariance, and finally (lower-right) the conditioned result in the original domain with the prior covariances indicated with thinner lines.}\label{fig:ill_rcud}
\end{figure}

\subsection{Inter-agent range measurement updates}\label{subsec:ranging}

Similar to the range constraint between the feet of an individual agent, the geometry of the inter-agent ranging gives the constraints
\begin{equation}\label{eq:range_without_tracked_points}
r-(\gamma_a+\gamma_b)\leq\|\mathbf{x}_a-\mathbf{x}_b\|\leq r+(\gamma_a+\gamma_b)
\end{equation}
where $r$ is the (true) range between agents' ranging devices and $\gamma_a$ and $\gamma_b$ are the maximum spatial separation of respective ranging device and $\mathbf{x}_a$ and $\mathbf{x}_b$; where in this case $\mathbf{x}_a$ and $\mathbf{x}_b$ are the positions of a foot of each agent.
The range only being dependent on $\|\mathbf{x}_a-\mathbf{x}_b\|$ means that the state transformation $\mathbf{z}=\mathbf{T}_{\mathbf{1}}\mathbf{x}$ where $\mathbf{1}=[1,1,1]$ and $\mathbf{z}_1=\mathbf{x}_a-\mathbf{x}_b$, and the corresponding mean and covariance transformations as explained in the previous subsection, can be used to let us exploit the marginalization~\eqref{eq:marginalization}.

The inter-agent ranging gives measurements $\tilde{r}$ of the range $r$. As reviewed in Section~\ref{subsec:fusion}, the malign attributes of $\tilde{r}$ which we have to deal with are potentially heavy-tailed error distributions and non-stationary spatially correlated errors due to diffraction, multi-path, or similar. This can be done by using the model $\tilde{r}=r+v+v'$ where $v$ is a heavy-tailed error component and $v'$ is a uniformly distributed component intended to cover the assumed bounded correlated errors in a manner similar to that of \cite{Lee2011}. Combining the model with \eqref{eq:range_without_tracked_points} and the state transformation $\mathbf{T_1}$ gives the measurement model
\begin{equation}\label{eq:ranging_info}
\tilde{r}-v-\gamma_r\leq\|\mathbf{z}_1\|\leq \tilde{r}-v+\gamma_r.
\end{equation}
where $\gamma_r$ is chosen to cover the bounds in \eqref{eq:range_without_tracked_points}, the asynchrony between $\hat{\mathbf{x}}_a$ and $\hat{\mathbf{x}}_b$, and the correlated errors $v'$. In practice $\gamma_r$ will be a system parameter trading integrity for information.

To update the global state estimate with the the range measurement $\tilde{r}$, the state $\hat{\mathbf{z}}_1$ and covariance estimates $\mathbf{P}_{z_1}$ must be conditioned on $\tilde{r}$ via \eqref{eq:ranging_info}. Due to the stochastic term $v$, we cannot use hard constraints as with the feet of a single agent.
However, by assigning a uniform prior to the constraint in \eqref{eq:ranging_info}, the likelihood function of $\tilde{r}$ given $\hat{\mathbf{z}}_1$ becomes
\begin{equation}\label{eq:likelihood}
f(\tilde{r}|\mathbf{z}_1)=\mathcal{U}(-\gamma_r,\gamma_r)\ast\mathcal{V}(\|\hat{\mathbf{z}}_1\|-\tilde{r},\sigma)
\end{equation}
where $\mathcal{U}(-\gamma_r,\gamma_r)$ is a uniform distribution over the interval $[-\gamma_r,\gamma_r]$, $\mathcal{V}(\|\hat{\mathbf{z}}_1\|-\tilde{r},\sigma_r)$ is the distribution of $v$ with mean $\|\hat{\mathbf{z}}_1\|-\tilde{r}$ and some scale $\sigma_r$, and $\ast$ denotes convolution.
Then, with the assumed Gaussian prior $\mathbf{z}_1\sim\mathcal{N}(\hat{\mathbf{z}}_1,\mathbf{P}_{z_1})$, the conditional distribution of $\mathbf{z}_1$ given $\tilde{r}$, $\hat{\mathbf{z}}_1$ , and $\mathbf{P}_{z_1}$ is
\begin{equation}\label{eq:prior}
f(\mathbf{z}_1|\tilde{r})\propto f(\tilde{r}|\hat{\mathbf{z}}_1)\,\mathcal{N}(\hat{\mathbf{z}}_1,\mathbf{P}_{z_1}).
\end{equation}
Since $\mathbf{z}_1$ is low-dimensional, the conditional moments $\hat{\mathbf{z}}_{1|\tilde{r}}$ and $\mathbf{C}_{z_1|\tilde{r}}$ can be evaluated by sampling. With the marginalization \eqref{eq:marginalization} and the inverse transformation $\mathbf{T}_{\mathbf{1}}^{-1}$, this will give the conditional mean and covariance of $\mathbf{x}$.




Since the likelihood function~\eqref{eq:likelihood} is typically heavy-tailed, it cannot easily be described by a set of samples. However, since the prior is (assumed) Gaussian, the sampling of it can efficiently be implemented with the eigenvalue decomposition. 
With sample points $\mathbf{u}^{(i)}$ of the standard Gaussian distribution, the corresponding sample points of the prior is given by
\begin{equation*}
\mathbf{s}^{(i)}=\hat{\mathbf{z}}_1+\mathbf{Q}\boldsymbol{\Lambda}^{1/2}\mathbf{u}^{(i)}
\end{equation*}
where $\mathbf{P}_{z_1}=\mathbf{Q}\boldsymbol{\Lambda}\mathbf{Q}^\top$ is the eigenvalue decomposition of $\mathbf{P}_{z_1}$. With the sample points $\mathbf{s}^{(i)}$, the associated prior weights only become dependent on $\mathbf{u}^{(i)}$ (apart from normalization) since
\begin{equation*}
w^{(i)}_\text{pr}\sim e^{-\tfrac{1}{2}(\mathbf{s}^{(i)}-\hat{\mathbf{z}}_{1})^\top\mathbf{P}^{-1}_{z_1}(\mathbf{s}^{(i)}-\hat{\mathbf{z}}_{1})} =e^{-\tfrac{1}{2}\|\mathbf{u}^{(i)}\|^2}
\end{equation*}
and can therefore, be precalculated. Reweighting with the likelihood function, $\tilde{w}^{(i)}_\text{po}=w^{(i)}_\text{pr}\cdot f(\tilde{r}|\mathbf{s}^{(i)})$ and normalizing the weights $w^{(i)}_\text{po}=\tilde{w}^{(i)}_\text{po}\cdot(\sum\tilde{w}^{(i)}_\text{po})^{-1}$, with suitable chosen $\mathbf{u}^{(i)}$ the conditional moments can be approximated by
\begin{equation*}
\hat{\mathbf{z}}_{1|\tilde{r}}\approx\medop\sum_i w^{(i)}_\text{po}\mathbf{s}^{(i)}\quad\text{and}\quad\mathbf{C}_{z_1|\tilde{r}}\approx\medop\sum_i w^{(i)}_\text{po}\mathbf{s}^{(i)}(\mathbf{s}^{(i)})^\top.
\end{equation*}
Consequently, as long as the likelihood function can be efficiently evaluated, any likelihood function may be used. For analytical convenience, we have typically let $\mathcal{V}(\cdot,\cdot)$ be Cauchy-distributed giving the heavy-tailed likelihood function
\begin{equation}\label{eq:ranging_likelihood}
f(\tilde{r}|\mathbf{s}^{(i)})\sim \text{atan}\left(\tfrac{\tilde{r}-\|\mathbf{s}^{(i)}\|+\gamma_r}{\sigma_r}\right)-\text{atan}\left(\tfrac{\tilde{r}-\|\mathbf{s}^{(i)}\|-\gamma_r}{\sigma_r}\right).
\end{equation}
The sampling based range update with this likelihood function and $\mathbf{u}^{(i)}$ from a square sampling lattice is illustrated in Fig.~\ref{fig:ill_rud}. Potential more elaborate techniques for choosing the sample points can be found in~\cite{Huber2008}.

\begin{figure}[t]
\centering
{\resizebox{\linewidth}{!}{\includegraphics{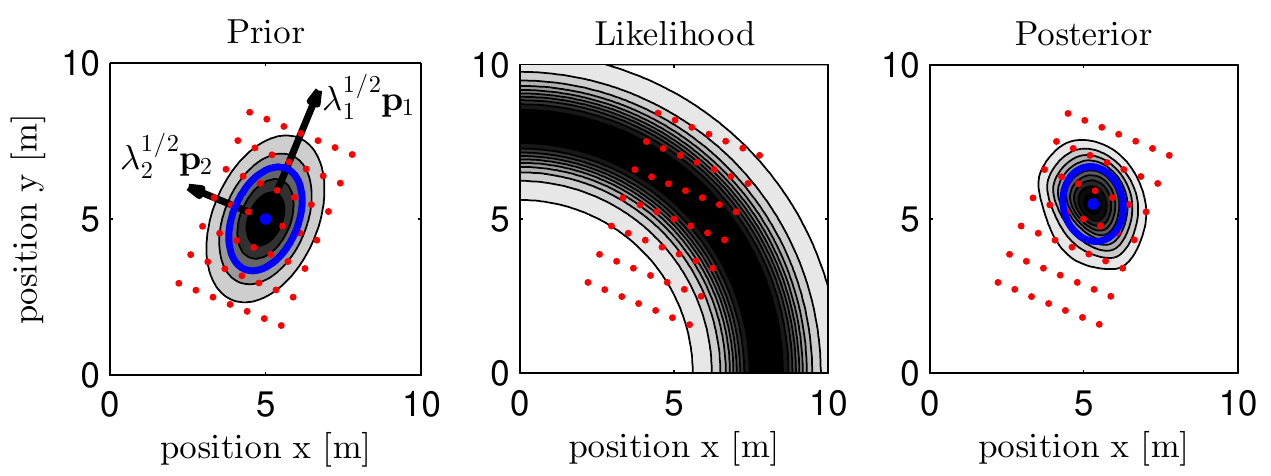}}}
\caption{Illustration of the suggested range update in two dimensions. From left to right, the plots show: the Gaussian prior given by the mean $\hat{\mathbf{z}}_1$ (blue dot) and the covariance $\mathbf{P}_{z_1}$ (blue ellipse) and with indicated samples (red dots) and eigenvectors/-values, the (one dimensional) likelihood function given by the range measurement $\tilde{r}$ and used to reweight the samples, and the resulting posterior with conditional mean $\hat{\mathbf{z}}_{1|r}$ and covariance $\mathbf{P}_{z_1|r}$ calculated from the reweighted samples.}\label{fig:ill_rud}
\end{figure}

The presented ranging update gives a robustness to outliers in the measurement data. In Fig.~\ref{fig:ill_rud2}, the influence functions for the sample based update and the traditional Kalman measurement update are shown for the ranging likelihood function \eqref{eq:ranging_likelihood} with $\gamma_r=2$ [m] and $\sigma_r=0.5$ [m] and position covariance values of $\mathbf{P}_{z_1}=\mathbf{I}$ [m$^2$] and $\mathbf{P}_{z_1}=0.3\,\mathbf{I}$ [m$^2$].
By comparing the blue solid and the red dashed-dotted lines, it is seen that when the position and ranging error covariances are of the same size, the suggested ranging update behaves like the Kalman update up to around three standard deviations, where it gracefully starts to neglect the range measurement. In addition, by comparing the blue dashed and the red dotted lines, it is seen that for smaller position error covariances, in contrast to the Kalman update, the suggested range update neglects ranging measurements with small errors (flat spot in the middle of the influence function). This has the effect that multiple ranging updates will not make the position error covariance collapse, which captures the fact that due to correlated errors, during standstill, multiple range measurements will contain a diminishing amount of information; and during motion the range measurements should only ``herd'' the dead reckoning. 

\begin{figure}[t]
\centering
{\resizebox{\linewidth}{!}{\includegraphics{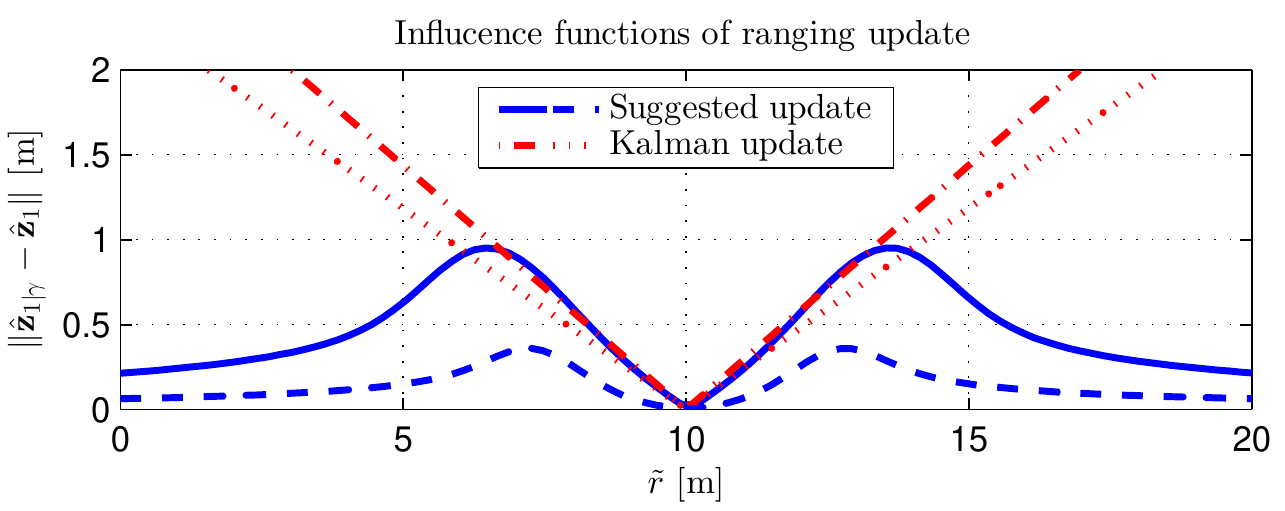}}}
\caption{Influence function of the ranging update for $\|\hat{\mathbf{z}}_1\|=10$ [m] performed with the suggested method (blue solid/dashed lines) and a traditional Kalman measurement update (red dashed-dotted/dotted lines) with $\mathbf{P}_{z_1}=\mathbf{I}$ [m$^2$] and $\mathbf{P}_{z_1}=0.3\mathbf{I}$ [m$^2$], respectively. For the suggested update $\gamma_r=2$ [m] and $v_2$ was Cauchy distributed with $\sigma_r=0.5$ [m] and for the Kalman measurement update the measurement error variance was 1 [m$^2$].
}\label{fig:ill_rud2}
\end{figure}



With slight modifications, the ranging updates can be used to incorporate information from many other information sources. Ranging to anchor nodes whose positions are not kept in $\mathbf{x}$ or position updates (from a GNSS receiver or similar) may trivially be implemented as range updates (zero range in the case of the position update) with $\mathbf{z}_1=\mathbf{x}_a-\mathbf{x}_b$ replaced with $\mathbf{z}_1=\mathbf{x}_a-\mathbf{x}_\text{c}$ where $\mathbf{x}_\text{c}$ is the position of the anchor node or the position measurement. Fusion of pressure measurements may be implemented as range updates in the vertical direction, either relative to other agents or relative to a reference pressure.

\subsection{Summary of sensor fusion}\label{subsec:sum_sf}
The central sensor fusion, as described in Section~\ref{sec:phys_arc}, keeps the position and heading of all feet in the global state vector $\mathbf{x}$. From all agents, it receives dead reckoning updates, $[d\mathbf{p}_\ell,d\psi_\ell]$, $\mathbf{P}_{\mathbf{p}_\ell}$, $\mathbf{P}_{\mathbf{p}_\ell,\psi_\ell}$, and $P_{\psi_\ell,\psi_\ell}$, and inter-agent range measurements $\tilde{r}$. The dead reckoning updates are used to propagate the corresponding states and covariances according to \eqref{eq:simple_ss}. At each dead reckoning update, the range constraint is imposed on the state space as described in subsection~\ref{subsec:constraint}, and corrections are sent back to the agent. The inter-agent range measurements are used to condition the state space as described in in subsection~\ref{subsec:ranging}. Pseudo-code for conditioning the state mean and covariance estimates on the range constraint and range measurements is shown in Algs.~\ref{Alg:range_constraint}-\ref{Alg:range_ud}.

\begin{algorithm}[t]
\caption{\footnotesize Pseudo-code for imposing the range constraint~\eqref{eq:range_constraint}, between navigation points $\mathbf{x}_a$ and $\mathbf{x}_b$, on the global state estimate $\hat{\mathbf{x}}$.}\label{Alg:range_constraint}
\begin{algorithmic}[1]
\STATE $\hat{\mathbf{z}}:=\mathbf{T}_\gamma\hat{\mathbf{x}}\quad\text{and}\quad\mathbf{P}_{z}:=\mathbf{T}_\gamma\mathbf{P}_{x}\mathbf{T}_\gamma^\top$
\STATE $\mathbf{L}:=\text{chol}(\mathbf{P}_{z_1})$
\STATE {\small Sample and project sigma points/weights} \\ $(\mathbf{z}^{(i)}_1,w^{(i)})\leftarrow(\hat{\mathbf{z}}_1,\mathbf{L})$
\STATE $\hat{\mathbf{z}}_{1|\gamma}:=\medop\sum_{i}w^{i}\mathbf{z}_1^{(i)}\quad\text{and}\quad\mathbf{C}_{z_1|\gamma}:=\medop\sum_{i}w^{i}\mathbf{z}_1^{(i)}(\mathbf{z}_1^{(i)})^\top$
\STATE {\small Calculate conditional mean and covariance by marginalization}\\ $(\hat{\mathbf{z}}_{|\gamma},\mathbf{P}_{z|\gamma})\leftarrow(\hat{\mathbf{z}}_{1|\gamma},\hat{\mathbf{z}}_2,\mathbf{C}_{z_1|\gamma},\mathbf{P}_z)$
\STATE $\hat{\mathbf{x}}_{|\gamma}:=\mathbf{T}_{\gamma}^{-1}\hat{\mathbf{z}}_{|\gamma}\quad\text{and}\quad\mathbf{P}_{x|\gamma}:=\mathbf{T}_{\gamma}^{-1}\mathbf{P}_{z|\gamma}\mathbf{T}_{\gamma}^{-\top}$
\end{algorithmic}
\end{algorithm}

\begin{algorithm}[t]
\caption{\footnotesize Pseudo-code for conditioning the global state estimate $\hat{\mathbf{x}}$ on the range measurements~\eqref{eq:ranging_info} between navigation points $\mathbf{x}_a$ and $\mathbf{x}_b$.}\label{Alg:range_ud}
\begin{algorithmic}[1]
\STATE $\hat{\mathbf{z}}:=\mathbf{T}_\mathbf{1}\hat{\mathbf{x}}\quad\text{and}\quad\mathbf{P}_{z}:=\mathbf{T}_\mathbf{1}\mathbf{P}_{x}\mathbf{T}_\mathbf{1}^\top$
\STATE $(\mathbf{Q},\boldsymbol{\Lambda}):=\text{eig}(\mathbf{P}_{z_1})$
\STATE $\mathbf{s}^{(i)}:=\hat{\mathbf{z}}_1+\mathbf{Q}\boldsymbol{\Lambda}^{1/2}\mathbf{u}^{(i)}\quad\forall i$
\STATE $\tilde{w}^{(i)}_\text{po}:=w^{(i)}_\text{pr}\cdot f(\tilde{r}|\mathbf{s}^{(i)})\quad\forall i$
\STATE $w^{(i)}_\text{po}:=\tilde{w}^{(i)}_\text{po}\cdot(\sum\tilde{w}^{(i)}_\text{po})^{-1}\quad\forall i$
\STATE $\hat{\mathbf{z}}_{1|\tilde{r}}:=\medop\sum_i w^{(i)}_\text{po}\mathbf{s}^{(i)}\quad\text{and}\quad\mathbf{C}_{z_1|\tilde{r}}:=\medop\sum_i w^{(i)}_\text{po}\mathbf{s}^{(i)}(\mathbf{s}^{(i)})^\top$
\STATE {\small Calculate conditional mean and covariance by marginalization}\\ $(\hat{\mathbf{z}}_{|\tilde{r}},\mathbf{P}_{z|\tilde{r}})\leftarrow(\hat{\mathbf{z}}_{1|\tilde{r}},\hat{\mathbf{z}}_2,\mathbf{C}_{z_1|\tilde{r}},\mathbf{P}_z)$
\STATE $\hat{\mathbf{x}}_{|\tilde{r}}:=\mathbf{T}_{\mathbf{1}}^{-1}\hat{\mathbf{z}}_{|\tilde{r}}\quad\text{and}\quad\mathbf{P}_{x|\tilde{r}}:=\mathbf{T}_{\mathbf{1}}^{-1}\mathbf{P}_{z|\tilde{r}}\mathbf{T}_{\mathbf{1}}^{-\top}$
\end{algorithmic}
\end{algorithm}

%
%
%
%

\section{Experimental results}\label{sec:exp_results}
To demonstrate the characteristics of the sensor fusion presented in the previous section, in the following subsection we first show numerical simulations giving a quantitative description of the fusion. Subsequently, to demonstrate the practical feasibility of the suggested architecture and sensor fusion, a real-time localization system implementation is briefly presented. 

\subsection{Simulations}
The cooperative localization by foot-mounted inertial sensors and inter-agent ranging is nonlinear and the behavior of the system will be highly dependent on the trajectories. Therefore, we cannot give an analytical expression for the performance. Instead, to demonstrate the system characteristics, two extreme scenarios are simulated. For both scenarios the agents move with 1~[m] steps at 1~[Hz]. Gaussian errors with standard deviation $0.01$~[m] and 0.2[$^\circ$] were added to the step displacements and the heading changes, respectively, and heavy-tailed Cauchy distributed errors of scale 1~[m] where added to the range measurements. The ranging is done time-multiplexed in a Round-robin fashion at a total rate of 1~[Hz].

\subsubsection{Straight-line march}
$N$ agents are marching beside each other in straight lines with agent separation of 10~[m]. The straight line is the worst case scenario for the dead reckoning and the position errors will be dominated by errors induced by the heading errors.  In Fig.~\ref{fig:test_trj}, an example of the estimated trajectories of the right (blue) and left (green) feet are shown from 3 agents without any further information, with range constraints between the feet, and with range constraints and inter-agent ranging.
The absolute and relative root-mean-square-error (RMSE) as a function of the walked distance, and for different number of agents, are shown in Fig.~\ref{fig:E_vs_dist}. The relative errors are naturally bounded by the inter agent ranging. However, the heading RMSE grows linearly with time/distance, and therefore, the absolute position error is seen to grow with distance. Similar behavior can be observed in the experimental data in~\cite{Rantakokko2012a,Nilsson2012a}. 
Naturally, the heading error, and therefore, also the absolute position RMSE drops as $1/\sqrt{N}$ where $N$ is the number of agents. This is shown in Fig.~\ref{fig:E_vs_N}. We may also note that the position errors of different agents become strongly correlated. The correlation coefficients for two agents as a function of distance are shown in Fig.~\ref{fig:corr_coef}.

\begin{figure}[t]
\centering
{\resizebox{0.95\linewidth}{!}{\includegraphics{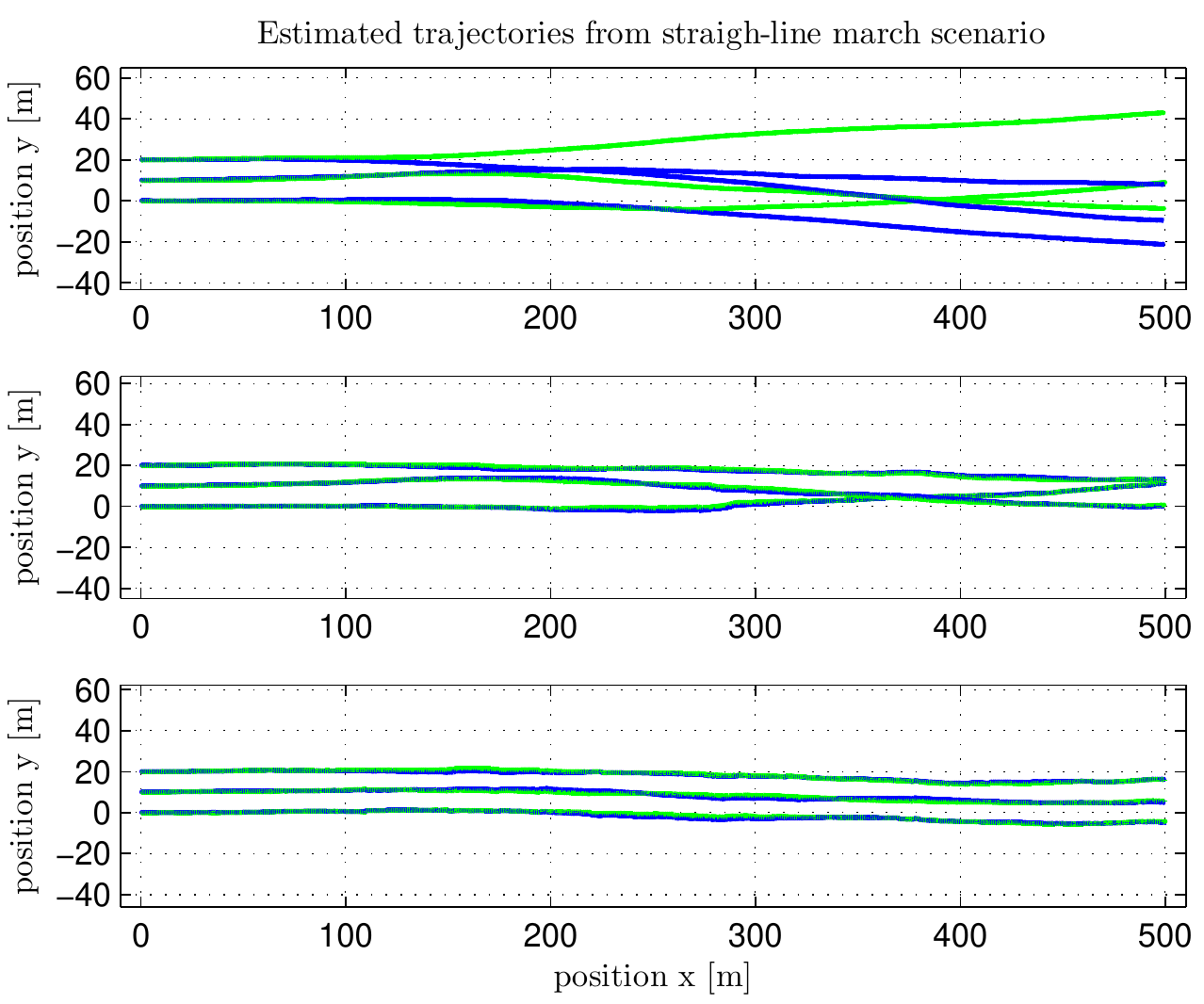}}}
\caption{Illustration of the gain of dual foot-mounted sensors and inter-agent ranging. The upper plot shows the step-wise dead reckoning of the individual feet (in blue and green) without any further information. The middle plot shows the step-wise dead reckoning with range constraints between the feet of individual agents. The lower plot shows the complete cooperative localization with step-wise dead reckoning, range constraints, and inter-agent ranging.}\label{fig:test_trj}
\end{figure}

\begin{figure}[t]
\centering
{\resizebox{0.95\linewidth}{!}{\includegraphics{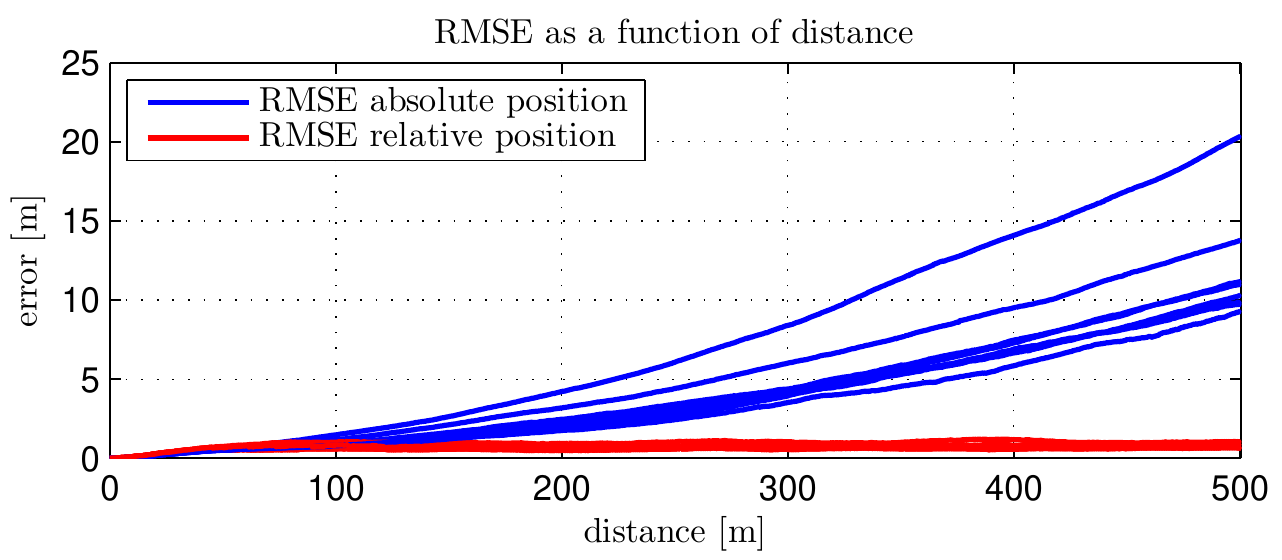}}}
\caption{Absolute position RMSE (blue lines) and relative position RMSE (red lines) as a function of distance for 100 Monte-Carlo runs. The different blue lines correspond, in ascending order, to increasing number of agents. Clearly, the relative error is bounded by the inter-agent ranging while the absolute error grows slower the larger the number of agents. The final position RMSEs as a function of the number of agents are shown in Fig.~\ref{fig:E_vs_N}.}\label{fig:E_vs_dist}
\end{figure}

\begin{figure}[t]
\centering
{\resizebox{0.95\linewidth}{!}{\includegraphics{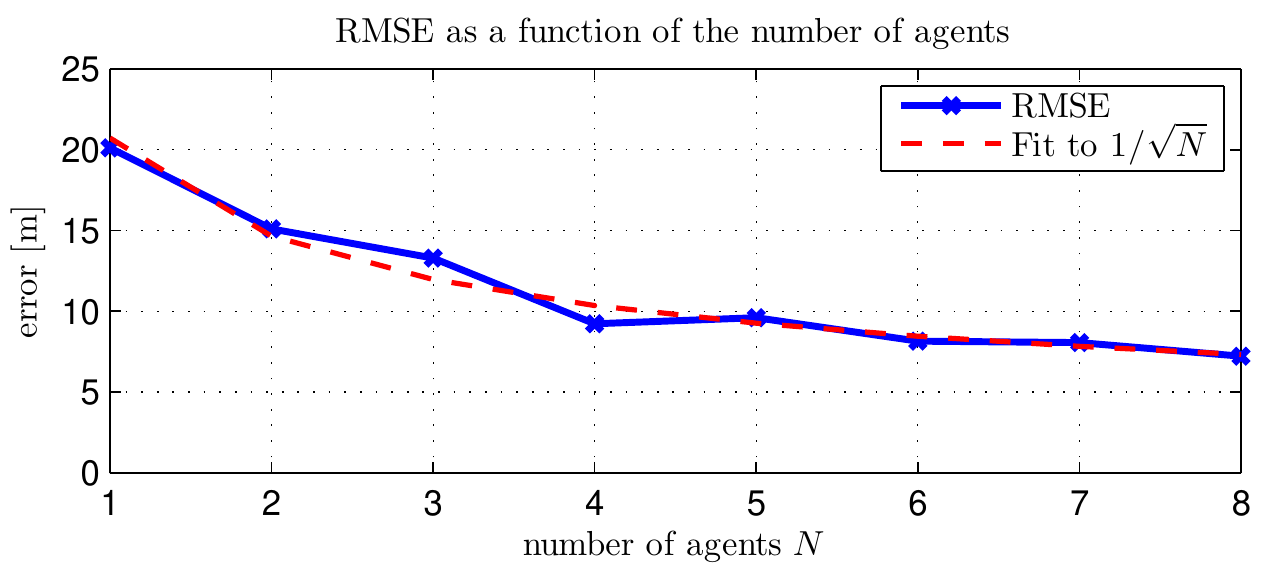}}}
\caption{Final position RMSE as a function of the number of agents (blue crossed line) for 100 Monte-Carlo runs. From the fit to $1/\sqrt{N}$ (dashed red line), the position error is seen to be decaying as the square-root of the number of agents.}\label{fig:E_vs_N}
\end{figure}

\begin{figure}[t]
\centering
{\resizebox{0.95\linewidth}{!}{\includegraphics{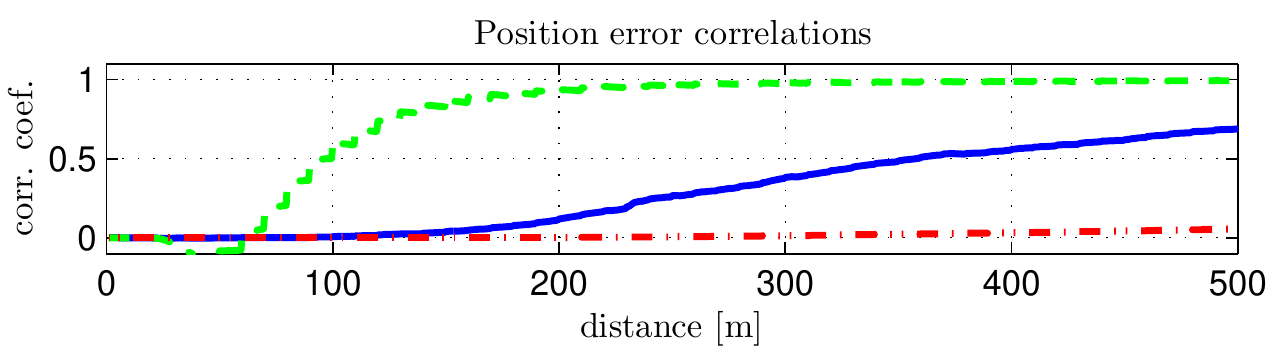}}}
\caption{Position error correlation coefficient for two agents in the straight-line march scenario in the x direction (solid blue), y direction (dashed green), and z direction (dotted/dashed red). Clearly, the positions errors of different agents become strongly correlated with increasing distance traveled.}\label{fig:corr_coef}
\end{figure}

\subsubsection{Three static agents}
Three non-collinear agents are standing still. This will be perceived by the foot-mounted inertial navigation, and therefore, they essentially become anchor nodes. This is obviously the best-case scenario. A fourth agent walk around them in a circle. An example of an estimated trajectory is shown in Fig.~\ref{fig:trj_static_agents} and the RMSE as a function of time is shown in Fig.~\ref{fig:E_static_agents}. Since anchor nodes are essentially present in the system, the errors are bounded. See~\cite{Kurazume1994} for further discussions. The non-zero RMSE reflects the range constraints in the system. 

\begin{figure}[t]
\centering
{\resizebox{0.95\linewidth}{!}{\includegraphics{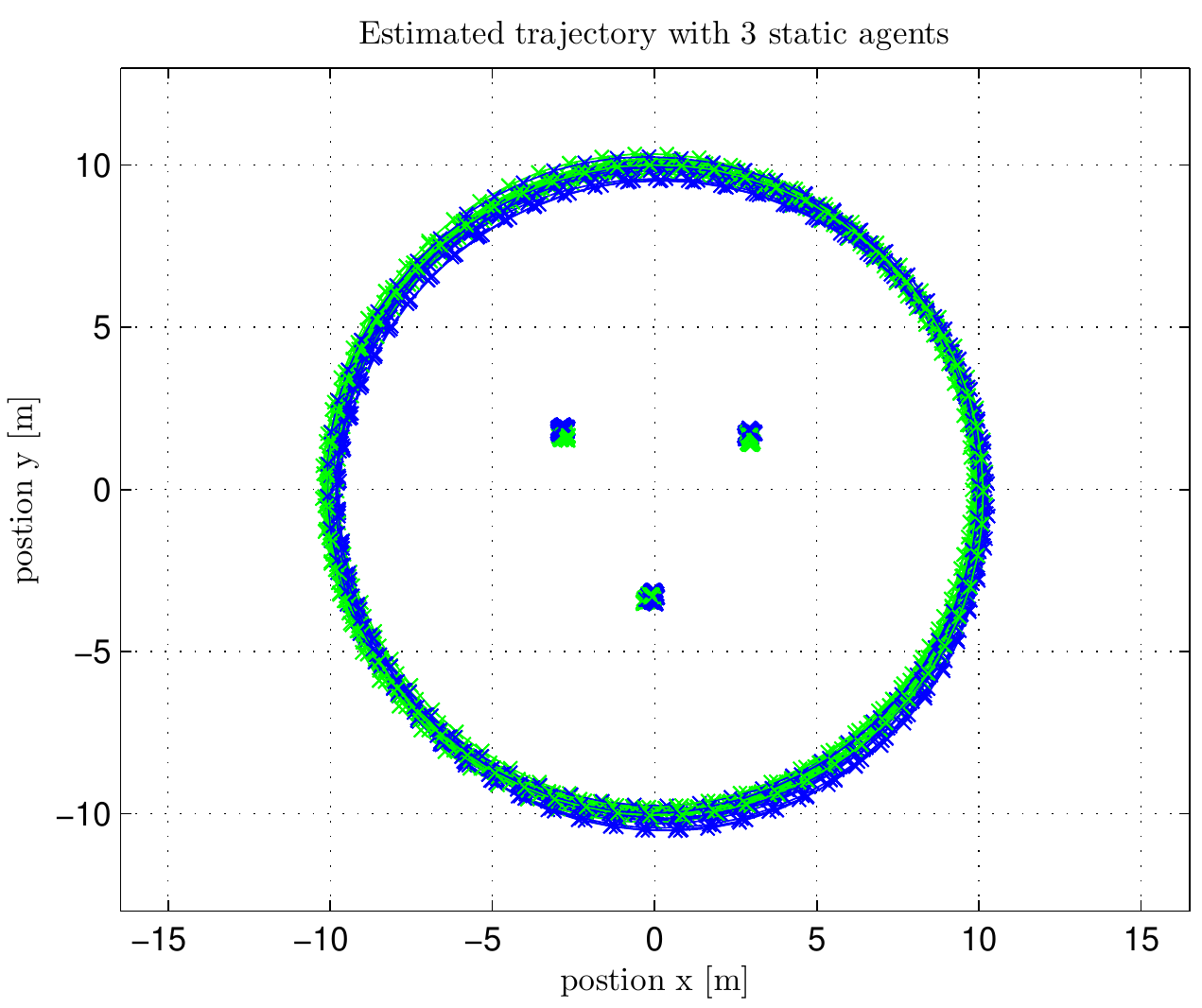}}}
\caption{The plot shows an estimated trajectory from the scenario with three static agents and a fourth mobile agent walking around them. Clearly, the position estimation errors are bounded.}\label{fig:trj_static_agents}
\end{figure}

\begin{figure}[t]
\centering
{\resizebox{0.99\linewidth}{!}{\includegraphics{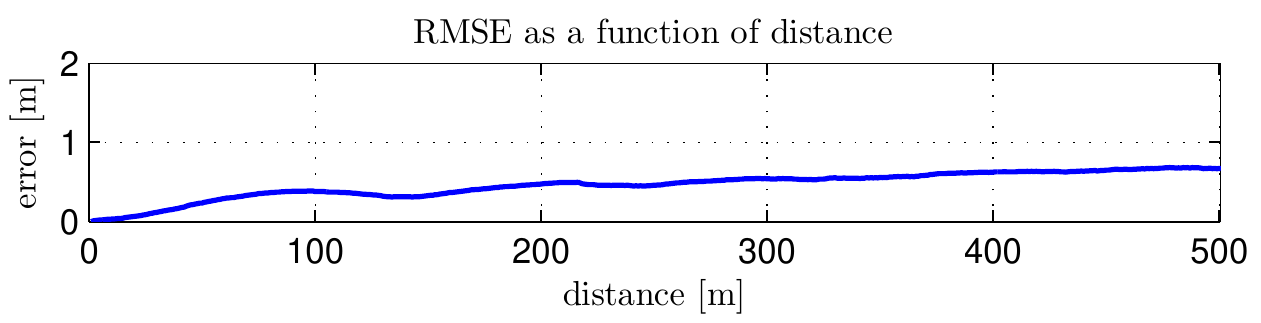}}}
\caption{The plot shows the RMSE of the mobile unit for the three static agents scenario over 100 Monter-Carlo runs. Being static, the three stationary agents essentially become anchor nodes, and therefore, the RMSE is bounded.}\label{fig:E_static_agents}
\end{figure}

From the two scenarios, we can conclude that the relative position errors are kept bounded by the inter-agent ranging while the absolute position errors (relative starting location) are bounded in the best-case (stationary agents); and that the error growth is reduced by a factor of $1/\sqrt{N}$ in the worst case. 


\subsection{Real-time implementation}
The decentralized system architecture has been realized with OpenShoe units ~\cite{Nilsson2012} and Android smart-phones and tablets (Samsung Galaxy S III and Tab 2 10.1) in the in-house developed tactical locator system TOR. The communication is done over an IEEE 802.11 WLAN. Synthetic inter-agent ranging is generated from position measurements from a Ubisense system (Ubisense Research \& Development Package), installed in the KTH R1 reactor hall~\cite{DeAngelis12}. The intension is to replace the Ubisense system with in-house developed UWB radios~\cite{Angelis2013}. The equipment for a single agent is shown in Fig.~\ref{fig:proto_equipment}. The multi-agent setup with additional equipment for sensor mounting is shown in Fig.~\ref{fig:multi_agent}.

\begin{figure}[t]
\centering
{\resizebox{\linewidth}{!}{\includegraphics{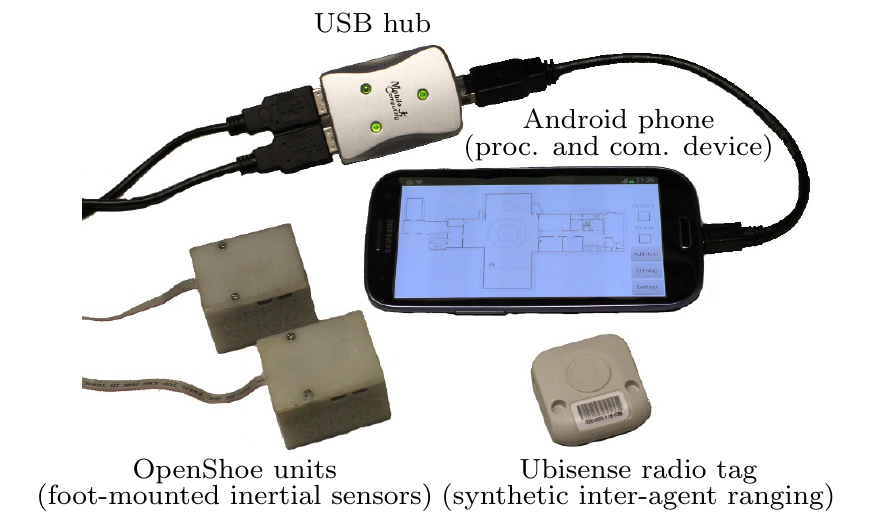}}}
\caption{Agent equipment carried by each agent in the prototype implementation. The OpenShoe units are connected to the USB-hub. Radio tags and a Ubisense real-time location system are used to generate synthetic range measurements between agents.}\label{fig:proto_equipment}
\end{figure}

\begin{figure}[t]
\centering
{\resizebox{\linewidth}{!}{\includegraphics{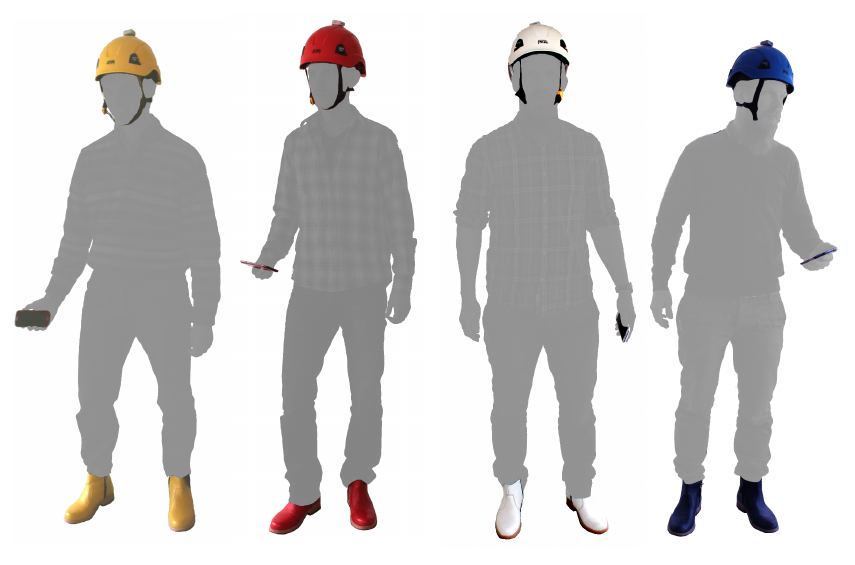}}}
\caption{Four agent with equipment displayed in Fig.~\ref{fig:proto_equipment}. The OpenShoe units are integrated in the soles of the shoes and the radio tags are attached to the helmets. The cables and the USB hubs are not displayed.}\label{fig:multi_agent}
\end{figure}

The step-wise inertial navigation and the associated transfer of displacements and heading changes has been implemented in the OpenShoe units. The agent filtering has been implemented as Android applications together with graphical user interfaces. A screen-shot of the graphical user interface with trajectories from a $\sim$10 [min] search in the reactor hall and adjacent rooms (built up walls not displayed) by 3 smoked divers
is shown in Fig.~\ref{fig:GUI}. The central sensor fusion has been implemented as a separate Android application running on one agent's Android platform.
Recently, voice radio communication and 3D audio has been integrated into the localization system~\cite{Nilsson2013}.

\begin{figure}[t]
\centering
{\resizebox{\linewidth}{!}{\includegraphics[trim=0mm 10mm 0mm 0mm,clip]{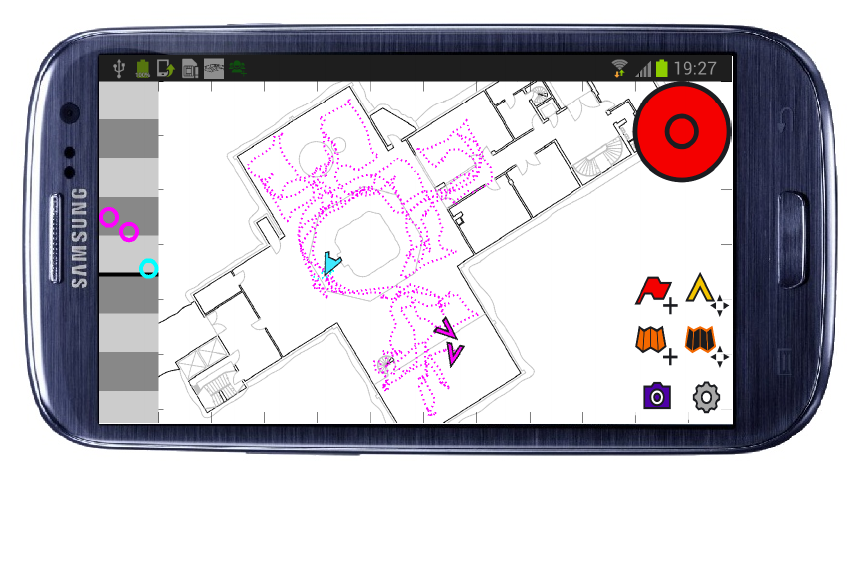}}}
\caption{Screen-shot of the cooperative localization system user interface overlayed on a photography of the Android phone. The interface shows the recursively estimated positions (trajectories) of 3 smoke divers, a smoke diving pair (magenta) and a smoke diving leader (blue), during a $\sim$10 minutes search of the R1 reactor hall and adjacent rooms. The built up rooms/walls are not displayed but can clearly be seen in the search pattern. At the time of the screen-shot, the smoke diving pair has advanced to the second floor as can be seen by the hight estimates displayed on the left side.}\label{fig:GUI}
\end{figure}

\section{Conclusions}\label{sec:Conclusions}
Key implementation challenges of cooperative localization by foot-mounted inertial sensors and inter-agent ranging are: designing an overall system architecture to minimize the required communication and computational cost, while retaining the performance and making it robust to varying connectivity; and fusing the information from the system under the constraint of the system architecture while retaining a high integrity and accuracy. A solution to the former problem has been presented in the partially decentralized system architecture based on the division and physical separation of the step-wise inertial navigation and the step-wise dead reckoning. A solution to the latter problem has been presented in the marginalization and sample based spatial separation constraint and ranging updates. By simulations, it has been shown that in the worst case scenario, the absolute localization RMSE improves as the square-root of the number of agents and the relative errors are bounded. In the best case scenario, both the relative and the absolute errors are bounded. Finally, the feasibility of the suggested architecture and sensor fusion has been demonstrated with simulations and a real-time system implementation featuring multiple agents and a meter level accuracy for operation times of tenth of minutes in a harsh industrial environment.


\ifdefined\IEEEVersion
{\footnotesize
\bibliographystyle{ieeetr}
\bibliography{coop}
}
\fi

\ifdefined\springerVersion
\bigskip


\section*{Competing interests}
The authors have no connection to any company whose products are referenced in the article. The authors declare that they have no competing interests.

\section*{Acknowledgements}
 Parts of this work have been fonded by the Swedish Agency for Innovation Systems (VINNOVA).


\newpage
{\ifthenelse{\boolean{publ}}{\footnotesize}{\small}
 \bibliographystyle{bmc_article}  
  \bibliography{coop} }     

\begin{thebibliography}{100}

\bibitem{Rantakokko2011}
J.~Rantakokko, J.~Rydell, P.~Str\"{o}mb\"{a}ck, P.~H\"{a}ndel, J.~Callmer,
  D.~T\"{o}rnqvist, F.~Gustafsson, M.~Jobs, and M.~Grud\'{e}n, ``Accurate and
  reliable soldier and first responder indoor positioning: multisensor systems
  and cooperative localization,'' {\em Wireless Communications, IEEE}, vol.~18,
  pp.~10 --18, april 2011.

\bibitem{Fuchs2011}
C.~Fuchs, N.~Aschenbruck, P.~Martini, and M.~Wieneke, ``Indoor tracking for
  mission critical scenarios: a survey,'' {\em Pervasive and Mobile Computing},
  vol.~7, no.~1, pp.~1 -- 15, 2011.

\bibitem{Pahlavan2002}
K.~Pahlavan, X.~Li, and J.~Makela, ``Indoor geolocation science and
  technology,'' {\em Communications Magazine, IEEE}, vol.~40, pp.~112 --118,
  feb 2002.

\bibitem{Renaudin2007}
V.~Renaudin, O.~Yalak, T.~P., and B.~Merminod, ``Indoor navigation of emergency
  agents,'' {\em European Journal of Navigation,}, vol.~5, pp.~36--42, 2007.

\bibitem{Glanzer2012}
G.~Glanzer, ``Personal and first-responder positioning: state of the art and
  future trends,'' in {\em Ubiquitous Positioning, Indoor Navigation, and
  Location Based Service (UPINLBS), 2012}, (Helsinki, Finland), 3-4 Oct. 2012.

\bibitem{Fischer2010}
C.~Fischer and H.~Gellersen, ``Location and navigation support for emergency
  responders: A survey,'' {\em Pervasive Computing, IEEE}, vol.~9, no.~1,
  pp.~38--47, 2010.

\bibitem{Hightower2001}
J.~Hightower and G.~Borriello, ``Location systems for ubiquitous computing,''
  {\em Computer}, vol.~34, no.~8, pp.~57--66, 2001.

\bibitem{Angermann2008}
M.~Angermann, M.~Khider, and P.~Robertson, ``Towards operational systems for
  continuous navigation of rescue teams,'' in {\em Position, Location and
  Navigation Symposium (PLANS), 2008 IEEE/ION}, pp.~153--158, 2008.

\bibitem{Mourikis2006}
A.~Mourikis and S.~Roumeliotis, ``On the treatment of relative-pose
  measurements for mobile robot localization,'' in {\em Robotics and
  Automation, 2006. ICRA 2006. Proceedings 2006 IEEE International Conference
  on}, (Orlando, FL, US), 15-19 May 2006.

\bibitem{Harle2013}
R.~Harle, ``A survey of indoor inertial positioning systems for pedestrians,''
  {\em Communications Surveys Tutorials, IEEE}, vol.~PP, no.~99, pp.~1--13,
  2013.

\bibitem{Foxlin2005}
E.~Foxlin, ``Pedestrian tracking with shoe-mounted inertial sensors,'' {\em
  Computer Graphics and Applications, IEEE}, vol.~25, pp.~38 --46, 2005.

\bibitem{Rantakokko2012}
J.~Rantakokko, E.~Emilsson, P.~Str\"{o}mb\"{a}ck, and J.~Rydell,
  ``Scenario-based evaluations of high-accuracy personal positioning systems,''
  in {\em Position, Location and Navigation Symposium (PLANS), 2012 IEEE/ION},
  (Myrtle Beach, SC, USA), 23-26 Apr. 2012.

\bibitem{Godha2006}
S.~Godha, G.~Lachapelle, and M.~E. Cannon, ``Integrated {GPS/INS} system for
  pedestrian navigation in a signal degraded environment,'' in {\em ION GNSS},
  (Fort Worth, TX, US), 26-29 Sept. 2006.

\bibitem{Laverne2011}
M.~Laverne, M.~George, D.~Lord, A.~Kelly, and T.~Mukherjee, ``Experimental
  validation of foot to foot range measurements in pedestrian tracking,'' in
  {\em ION GNSS}, (Portland, OR, US), 19-23 Sept. 2011.

\bibitem{Nilsson2012}
J.-O. Nilsson, I.~Skog, P.~H\"{a}ndel, and K.~Hari, ``Foot-mounted inertial
  navigation for everybody -- {A}n open-source embedded implementation,'' in
  {\em Position, Location and Navigation Symposium (PLANS), 2012 IEEE/ION},
  (Myrtle Beach, SC, USA), 23-26 Apr. 2012.

\bibitem{Fischer2012}
C.~Fischer, P.~Talkad~Sukumar, and M.~Hazas, ``Tutorial: Implementing a
  pedestrian tracker using inertial sensors,'' {\em Pervasive Computing, IEEE},
  vol.~12, no.~2, pp.~17--27, 2013.

\bibitem{instk}
``{Open Source Aided Inertial Navigation ToolKit}.''
\newblock (available at: http://www.instk.org).

\bibitem{InterSense}
``Navshoe$^\text{\textsc{tm}}$.''
\newblock Product by {I}nter{S}ense {I}nc., (http://www.intersense.com).

\bibitem{aionav}
``{AIONAV-F/M/P/L}.''
\newblock Products by {AIONAV Systems Ltd} (http://www.aionav.com).

\bibitem{MINT}
``Micro-inertial navigation technology ({MINT}).''
\newblock Technology by Carnegie Mellon, The Robotics Institute, National
  Robotics Engineering Center ({NREC}),
  (http://www.rec.ri.cmu.edu/projects/mint/).

\bibitem{WPI}
``Precision indoor personnel location \& tracking annual international
  technology workshop,'' WPI (Worcester Polytechnic Institute), Worcester,
  Massachusetts, 2006 -- 2012.
\newblock Workshop series, Retrieved at:
  http://www.wpi.edu/academics/ece/ppl/workshops.html.

\bibitem{Gezici2005}
S.~Gezici, Z.~Tian, G.~Giannakis, H.~Kobayashi, A.~Molisch, H.~Poor, and
  Z.~Sahinoglu, ``Localization via ultra-wideband radios: a look at positioning
  aspects for future sensor networks,'' {\em Signal Processing Magazine, IEEE},
  vol.~22, pp.~70 -- 84, july 2005.

\bibitem{Patwari2005}
N.~Patwari, J.~Ash, S.~Kyperountas, A.~Hero~III, R.~Moses, and N.~Correal,
  ``Locating the nodes: cooperative localization in wireless sensor networks,''
  {\em Signal Processing Magazine, IEEE}, vol.~22, pp.~54 -- 69, july 2005.

\bibitem{Win2011}
M.~Win, A.~Conti, S.~Mazuelas, Y.~Shen, W.~Gifford, D.~Dardari, and M.~Chiani,
  ``Network localization and navigation via cooperation,'' {\em Communications
  Magazine, IEEE}, vol.~49, pp.~56 --62, may 2011.

\bibitem{Wymeersch2009}
H.~Wymeersch, J.~Lien, and M.~Win, ``Cooperative localization in wireless
  networks,'' {\em Proceedings of the IEEE}, vol.~97, pp.~427 --450, feb. 2009.

\bibitem{Angelis2013}
A.~{De Angelis}, S.~Dwivedi, and P.~H\"{a}ndel, ``Characterization of a
  flexible {UWB} sensor for indoor localization,'' {\em Instrumentation and
  Measurement, IEEE Transactions on}, vol.~62, no.~5, pp.~905--913, 2013.

\bibitem{Alessio2009}
A.~{De Angelis}, M.~Dionigi, A.~Moschitta, and P.~Carbone, ``A low-cost
  ultra-wideband indoor ranging system,'' {\em Instrumentation and Measurement,
  IEEE Transactions on}, vol.~58, pp.~3935 --3942, dec. 2009.

\bibitem{Moragrega2009}
A.~Moragrega, X.~Artiga, C.~Gavrincea, C.~Ibars, M.~Navarro, M.~Najar,
  P.~Miskovsky, F.~Mira, and M.~di~Renzo, ``Ultra-wideband testbed for 6.0-8.5
  {GHz} ranging and low data rate communication,'' in {\em Radar Conference,
  2009. EuRAD 2009. European}, (Rome, Italy), 30 Sept. - 2 Oct. 2009.

\bibitem{Karbownik2012}
P.~Karbownik, G.~Krukar, M.~M. Pietrzyk, N.~Franke, and T.~von~der Gr\"{u}n,
  ``Ultra-wideband technology-based localization platform -- architecture \&
  experimental validation,'' in {\em Indoor Positioning and Indoor Navigation
  (IPIN), 2012 International Conference on}, (Sydney, Australia), 13-15 Nov.
  2012.

\bibitem{Cazzorla2013}
A.~Cazzorla, G.~De~Angelis, A.~Moschitta, M.~Dionigi, F.~Alimenti, and
  P.~Carbone, ``A 5.6-{GHz UWB} position measurement system,'' {\em
  Instrumentation and Measurement, IEEE Transactions on}, vol.~62, no.~3,
  pp.~675--683, 2013.

\bibitem{TimeDomain}
``{PulsOn}$^\text{\textsc{tm}}$.''
\newblock Product by Time Domain, (http://www.timedomain.com).

\bibitem{Ensco1}
B.~D. Farnsworth and D.~W. Taylor, ``High-precision 2.4 {GHz DSSS RF}
  ranging.'' White Paper, Ensco Inc.
\newblock (http://www.ensco.com).

\bibitem{nanotron}
Product range by {Nanotron Technologies}, (http://www.nanotron.com).

\bibitem{Motorola}
``{XIR P8660/APX 7000}.''
\newblock Product by Motorola, Inc. (http://www.motorola.com).

\bibitem{Interspiro}
``{SpiroLink}.''
\newblock Product by Interspiro, (http://www.interspiro.com).

\bibitem{Thales}
``{AN/PRC-148 MBITR/JEM and AN/PRC-154 Rifleman Radio}.''
\newblock Products by Thales Communications, Inc.
  (http://www.thalescomminc.com).

\bibitem{Zephyr}
``{Tactical Radio Comms}.''
\newblock Product by Zephyr Technology Corp.
  (http://www.zephyr-technology.com).

\bibitem{Stromback2010}
P.~Str\"{o}mb\"{a}ck, J.~Rantakokko, S.-L. Wirkander, M.~Alexandersson,
  K.~Fors, I.~Skog, and P.~H\"{a}ndel, ``Foot-mounted inertial navigation and
  cooperative sensor fusion for indoor positioning,'' in {\em ION International
  Technical Meeting (ITM)}, (San Diego, CA, US), 25-27 Jan. 2010.

\bibitem{Hawkinson2012}
W.~Hawkinson, P.~Samanant, R.~McCroskey, R.~Ingvalson, A.~Kulkarni, L.~Haas,
  and B.~English, ``{GLANSER: Geospatial location, accountability, and
  Navigation System for Emergency Responders} - system concept and performance
  assessment,'' in {\em Position, Location and Navigation Symposium (PLANS),
  2012 IEEE/ION}, (Myrtle Beach, SC, US), 24-26 Apr. 2012.

\bibitem{Kloch2011}
K.~Kloch, P.~Lukowicz, and C.~Fischer, ``Collaborative {PDR} localisation with
  mobile phones,'' in {\em Wearable Computers (ISWC), 2011 15th Annual
  International Symposium on}, (San Francisco, CA, US), 12-15 Jun. 2011.

\bibitem{Kamisaka2012}
D.~Kamisaka, T.~Watanabe, S.~Muramatsu, A.~Kobayashi, and H.~Yokoyama,
  ``Estimating position relation between two pedestrians using mobile phones,''
  in {\em Pervasive Computing}, vol.~7319 of {\em Lecture Notes in Computer
  Science}, pp.~307--324, Springer Berlin Heidelberg, 2012.

\bibitem{Harris2013}
M.~Harris, ``The way through the flames,'' {\em IEEE Spectrum}, vol.~9,
  pp.~26--31, 2013.

\bibitem{Adhoc}
``Topics in ad hoc and sensor networks,'' {\em Series of the IEEE Communication
  Magazine}, 2005 -- present.

\bibitem{adhocnetworks}
``{Ad Hoc Networks. Elsevier journal},'' 2003 -- present.

\bibitem{Bruno2005}
R.~Bruno, M.~Conti, and E.~Gregori, ``Mesh networks: commodity multihop ad hoc
  networks,'' {\em Communications Magazine, IEEE}, vol.~43, pp.~123 -- 131,
  march 2005.

\bibitem{Blazevic2005}
L.~Blazevic, J.-Y. Le~Boudec, and S.~Giordano, ``A location-based routing
  method for mobile ad hoc networks,'' {\em Mobile Computing, IEEE Transactions
  on}, vol.~4, pp.~97 -- 110, march-april 2005.

\bibitem{Mauve2001}
M.~Mauve, J.~Widmer, and H.~Hartenstein, ``A survey on position-based routing
  in mobile ad hoc networks,'' {\em Network, IEEE}, vol.~15, pp.~30 --39,
  nov.-dec. 2001.

\bibitem{Nilsson2013a}
J.-O. Nilsson and P.~H\"{a}ndel, ``Recursive {Bayesian} initialization of
  localization based on ranging and dead reckoning,'' in {\em Intelligent
  Robots and Systems, 2013. IROS 2013. IEEE/RSJ International Conference on},
  (Tokyo, Japan), 3-7 Nov. 2013.

\bibitem{Nilsson2010a}
J.-O. Nilsson and P.~H\"{a}ndel, ``Time synchronization and temporal ordering
  of asynchronous sensor measurements of a multi-sensor navigation system,'' in
  {\em Position, Location and Navigation Symposium (PLANS), 2010 IEEE/ION},
  (Palm Springs, CA, US), 3-6 May 2010.

\bibitem{Kschischang2001}
F.~Kschischang, B.~Frey, and H.-A. Loeliger, ``Factor graphs and the
  sum-product algorithm,'' {\em Information Theory, IEEE Transactions on},
  vol.~47, pp.~498 --519, feb 2001.

\bibitem{Ihler2005}
A.~Ihler, I.~Fisher, J.W., R.~Moses, and A.~Willsky, ``Nonparametric belief
  propagation for self-localization of sensor networks,'' {\em Selected Areas
  in Communications, IEEE Journal on}, vol.~23, pp.~809 -- 819, april 2005.

\bibitem{Mazuelas2011}
S.~Mazuelas, Y.~Shen, and M.~Win, ``Information coupling in cooperative
  localization,'' {\em Communications Letters, IEEE}, vol.~15, pp.~737 --739,
  july 2011.

\bibitem{Bahr2009a}
A.~Bahr, M.~Walter, and J.~Leonard, ``Consistent cooperative localization,'' in
  {\em Robotics and Automation, 2009. ICRA '09. IEEE International Conference
  on}, (Kobe, Japan), 12-17 May 2009.

\bibitem{Nerurkar2010}
E.~Nerurkar and S.~Roumeliotis, ``Asynchronous multi-centralized cooperative
  localization,'' in {\em Intelligent Robots and Systems (IROS), 2010 IEEE/RSJ
  International Conference on}, (Taipei, Taiwan), 18-22 Oct. 2010.

\bibitem{Romanovas2012}
M.~Romanovas, V.~Goridko, A.~Al-Jawad, M.~Schwaab, L.~Klingbeil, M.~Traechtler,
  and Y.~Manoli, ``A study on indoor pedestrian localization algorithms with
  foot-mounted sensors,'' in {\em Indoor Positioning and Indoor Navigation
  (IPIN), 2012 International Conference on}, (Sydney, Australia), 13-15 Nov.
  2012.

\bibitem{Gadeke2012}
T.~Gadeke, J.~Schmid, M.~Zahnlecker, W.~Stork, and K.~D. Muller-Glaser,
  ``Smartphone pedestrian navigation by foot-{IMU} sensor fusion,'' in {\em
  Ubiquitous Positioning, Indoor Navigation, and Location Based Service
  (UPINLBS), 2012}, (Helsinki, Finland), 3-4 Oct. 2012.

\bibitem{Roumeliotis2002}
S.~Roumeliotis and G.~Bekey, ``Distributed multirobot localization,'' {\em
  Robotics and Automation, IEEE Transactions on}, vol.~18, pp.~781 -- 795, oct
  2002.

\bibitem{Martinelli2007}
A.~Martinelli, ``Improving the precision on multi robot localization by using a
  series of filters hierarchically distributed,'' in {\em Intelligent Robots
  and Systems, 2007. IROS 2007. IEEE/RSJ International Conference on}, (San
  Diego, CA, US), 29 Sept.- 2 Nov. 2007.

\bibitem{Bahr2009}
A.~Bahr, J.~J. Leonard, and M.~F. Fallon, ``Cooperative localization for
  autonomous underwater vehicles,'' {\em The International Journal of Robotics
  Research}, vol.~28, pp.~714--728, 2009.

\bibitem{Krach2008a}
B.~Krach and P.~Robertson, ``Integration of foot-mounted inertial sensors into
  a {Bayesian} location estimation framework,'' in {\em Positioning, Navigation
  and Communication, 2008. WPNC 2008. 5th Workshop on}, (Hannover, Germany), 27
  Mar. 2008.

\bibitem{Angermann2012}
M.~Angermann and P.~Robertson, ``{FootSLAM}: {P}edestrian simultaneous
  localization and mapping without exteroceptive sensors -- {H}itchhiking on
  human perception and cognition,'' {\em Proceedings of the IEEE}, vol.~100,
  pp.~1840--1848, 2012.

\bibitem{Roberston2010}
P.~Robertson, M.~Angermann, B.~Krach, and M.~Khider, ``{SLAM Dance} --
  inertial-based joint mapping and positioning for pedestrian navigation,''
  {\em Inside GNSS}, vol.~5, May 2010.

\bibitem{Pinchin2012}
J.~Pinchin, C.~Hide, and T.~Moore, ``A particle filter approach to indoor
  navigation using a foot mounted inertial navigation system and heuristic
  heading information,'' in {\em Indoor Positioning and Indoor Navigation
  (IPIN), 2012 International Conference on}, (Sydney, Australia), 13-15 Nov.
  2012.

\bibitem{Rantakokko2012a}
J.~Rantakokko, P.~Str\"{o}mb\"{a}ck, E.~Emilsson, and J.~Rydell, ``Soldier
  positioning in {GNSS}-denied operations,'' in {\em {Navigation Sensors and
  Systems in GNSS Denied Environments (SET-168)}}, (Izmir, Turkey), 8-9 Oct.
  2012.

\bibitem{Godha2008}
S.~Godha and G.~Lachapelle, ``Foot mounted inertial system for pedestrian
  navigation,'' {\em Measurement Science and Technology}, vol.~19, no.~7, 2008.

\bibitem{Bird2011}
J.~Bird and D.~Arden, ``Indoor navigation with foot-mounted strapdown inertial
  navigation and magnetic sensors [emerging opportunities for localization and
  tracking],'' {\em Wireless Communications, IEEE}, vol.~18, pp.~28 --35, april
  2011.

\bibitem{Bancroft2008}
J.~B. Bancroft, G.~Lachapelle, M.~E. Cannon, and M.~G. Petovello, ``Twin
  {IMU-HSGPS} integration for pedestrian navigation,'' in {\em ION GNSS},
  (Savannah, GA, US), 15-16 Sept. 2008.

\bibitem{Brand2003}
T.~Brand and R.~Phillips, ``Foot-to-foot range measurement as an aid to
  personal navigation,'' in {\em Proc. 59th Annual Meeting of The Institute of
  Navigation and CIGTF 22nd Guidance Test Symposium}, (Albuquerque, NM, US),
  23-25 Jun. 2003.

\bibitem{Hung2013}
T.~N. Hung and Y.~S. Suh, ``Inertial sensor-based two feet motion tracking for
  gait analysis,'' {\em Sensors}, vol.~13, pp.~5614 -- 5629, april 2013.

\bibitem{Saarinen2004}
J.~Saarinen, J.~Suomela, S.~Heikkila, M.~Elomaa, and A.~Halme, ``Personal
  navigation system,'' in {\em Intelligent Robots and Systems, 2004. (IROS
  2004). Proceedings. 2004 IEEE/RSJ International Conference on}, (Sendai,
  Japan), 28 Sept.-2 oct. 2004.

\bibitem{Zhang2007}
Y.~P. Zhang, L.~Bin, and C.~Qi, ``Characterization of on-human body {UWB} radio
  propagation channel,'' {\em Microwave and Optical Technology Letters},
  vol.~49, no.~6, pp.~1365--1371, 2007.

\bibitem{Zachariah2012}
D.~Zachariah, I.~Skog, M.~Jansson, and P.~H\"{a}ndel, ``Bayesian estimation
  with distance bounds,'' {\em Signal Processing Letters, IEEE}, vol.~19,
  pp.~880 --883, dec. 2012.

\bibitem{Skog2012}
I.~Skog, J.-O. Nilsson, D.~Zachariah, and P.~H\"{a}ndel, ``Fusing information
  from two navigation system using an upper bound on their maximum spatial
  separation,'' in {\em Indoor Positioning and Indoor Navigation (IPIN), 2012
  International Conference on}, (Sydney, Australia), 13-15 Nov. 2012.

\bibitem{Prateek2013}
G.~V. Prateek, R.~Girisha, K.~Hari, and P.~H\"{a}ndel, ``Data fusion of dual
  foot-mounted {INS} to reduce the systematic heading drift,'' in {\em Proc.
  4th International Conference on Intelligent Systems, Modelling and
  Simulation}, (Bangkok, Thailand), 29-31 Jan. 2013.

\bibitem{Lee2011}
S.~Lee, B.~Kim, H.~Kim, R.~Ha, and H.~Cha, ``Inertial sensor-based indoor
  pedestrian localization with minimum 802.15.4a configuration,'' {\em
  Industrial Informatics, IEEE Transactions on}, vol.~7, pp.~455 --466, aug.
  2011.

\bibitem{Abrudan2011}
T.~E. Abrudan, L.~M. Paula, J.~{a}o Barros, J.~{a}o Paulo Silva~Cunha, and
  N.~B. de~Carvalho, ``Indoor location estimation and tracking in wireless
  sensor networks using a dual frequency approach,'' in {\em Indoor Positioning
  and Indoor Navigation (IPIN), 2011 International Conference on},
  (Guimar\~{a}es, Portugal), 21-23 Sept. 2011.

\bibitem{Hashemi1993}
H.~Hashemi, ``The indoor radio propagation channel,'' {\em Proceedings of the
  IEEE}, vol.~81, pp.~943 --968, jul 1993.

\bibitem{Pahlavan2006}
K.~Pahlavan, F.~O. Akgul, M.~Heidari, A.~Hatami, J.~M. Elwell, and R.~D.
  Tingley, ``Indoor geolocation in the absence of direct path,'' {\em Wireless
  Communications, IEEE}, vol.~13, no.~6, pp.~50 --58, 2006.

\bibitem{Yoon2011}
C.~Yoon and H.~Cha, ``Experimental analysis of {IEEE} 802.15.4a {CSS} ranging
  and its implications,'' {\em Computer Communications}, vol.~34, no.~11,
  pp.~1361 -- 1374, 2011.

\bibitem{Olson2006}
E.~Olson, J.~Leonard, and S.~Teller, ``Robust range-only beacon localization,''
  {\em Oceanic Engineering, IEEE Journal of}, vol.~31, no.~4, pp.~949--958,
  2006.

\bibitem{Whitehouse2006}
K.~Whitehouse and D.~Culler, ``A robustness analysis of multi-hop ranging-based
  localization approximations,'' in {\em Proceedings of the 5th international
  conference on Information processing in sensor networks}, (Nashville, TN,
  US), 19-21 Apr. 2006.

\bibitem{Zoubir2012}
A.~Zoubir, V.~Koivunen, Y.~Chakhchoukh, and M.~Muma, ``Robust estimation in
  signal processing: a tutorial-style treatment of fundamental concepts,'' {\em
  Signal Processing Magazine, IEEE}, vol.~29, pp.~61 --80, july 2012.

\bibitem{Zampella2012}
F.~Zampella, A.~D. Angelis, I.~Skog, D.~Zachariah, and A.~Jim\'{e}nez, ``A
  constraint approach for {UWB} and {PDR} fusion,'' in {\em Indoor Positioning
  and Indoor Navigation (IPIN), 2012 International Conference on}, (Sydney,
  Australia), 13-15 Nov. 2012.

\bibitem{Sornette2001}
D.~Sornette and K.~Ide, ``The {Kalman-L\'{e}vy} filter,'' {\em Physica D:
  Nonlinear Phenomena}, vol.~151, no.~2-4, pp.~142 -- 174, 2001.

\bibitem{Vila2011}
J.~Vil\`{a}-Valls, C.~Fern\'{a}ndez-Prades, P.~Closas, and J.~A.
  Fernandez-Rubio, ``Bayesian filtering for nonlinear state-space models in
  symmetric $\alpha$-stable measurement noise,'' in {\em Proc. 19th European
  Signal Processing Conference (EUSIPCO 2011)}, (Barcelona, Spain), 29 Aug.-2
  Sept. 2011.

\bibitem{Gordon2003}
N.~Gordon, J.~Percival, and M.~Robinson, ``The {Kalman-Levy} filter and
  heavy-tailed models for tracking manoeuvring targets,'' in {\em Information
  Fusion, 2003. Proceedings of the Sixth International Conference of}, (Cairns,
  Australia), 8-11 Jul. 2003.

\bibitem{Wang2008}
D.~Wang, C.~Zhang, and X.~Zhao, ``Multivariate {L}aplace filter: a heavy-tailed
  model for target tracking,'' in {\em Pattern Recognition, 2008. ICPR 2008.
  19th International Conference on}, (Tampa, FL, US), 8-11 Dec. 2008.

\bibitem{Allen2010}
R.~Allen, K.-C. Lin, and C.~Xu, ``Robust estimation of a maneuvering target
  from multiple unmanned air vehicles' measurements,'' in {\em Collaborative
  Technologies and Systems (CTS), 2010 International Symposium on}, (Chicago,
  IL, US), 17-21 May 2010.

\bibitem{Britting1971}
K.~R. Britting, {\em Inertial Navigation Systems Analysis}.
\newblock John Wiley \& Sons, Inc., 1971.

\bibitem{Jekeli2001}
C.~Jekeli, {\em Inertial Navigation Systems with Geodetic Applications}.
\newblock de Gruyter, 2001.

\bibitem{Nilsson2012a}
J.-O. Nilsson, I.~Skog, and P.~H\"{a}ndel, ``A note on the limitations of
  {ZUPTs} and the implications on sensor error modeling,'' in {\em Indoor
  Positioning and Indoor Navigation (IPIN), 2012 International Conference on},
  (Sydney, Australia), 13-15 Nov. 2012.

\bibitem{Kelly2011}
A.~Kelly, ``Personal navigation system based on dual shoe mounted {IMUs} and
  intershoe ranging,'' in {\em Precision Indoor Personnel Location \& Tracking
  Annual International Technology Workshop}, (Worcester, MA, US), 1 Aug. 2011.

\bibitem{Skog2010}
I.~Skog, P.~H\"{a}ndel, J.-O. Nilsson, and J.~Rantakokko, ``Zero-velocity
  detection: an algorithm evaluation,'' {\em Biomedical Engineering, IEEE
  Transactions on}, vol.~57, pp.~2657 --2666, nov. 2010.

\bibitem{Skog2010a}
I.~Skog, J.-O. Nilsson, and P.~H\"{a}ndel, ``Evaluation of zero-velocity
  detectors for foot-mounted inertial navigation systems,'' in {\em Indoor
  Positioning and Indoor Navigation (IPIN), 2010 International Conference on},
  (Z\"{u}rich, Switzerland), 15-17 Sept. 2010.

\bibitem{Nilsson2013b}
J.-O. Nilsson and P.~H\"{a}ndel, ``Standing still with inertial navigation,''
  in {\em Indoor Positioning and Indoor Navigation (IPIN), 2013 International
  Conference on}, (Montb\'{e}liard - Belfort, France), 28-31 Oct. 2013.

\bibitem{Bebek2010}
{\"{O}}.~Bebek, M.~Suster, S.~Rajgopal, M.~Fu, X.~Huang,
  M.~\c{C}avu\c{s}o\u{g}lu, D.~Young, M.~Mehregany, A.~van~den Bogert, and
  C.~Mastrangelo, ``Personal navigation via high-resolution gait-corrected
  inertial measurement units,'' {\em Instrumentation and Measurement, IEEE
  Transactions on}, vol.~59, pp.~3018 --3027, nov. 2010.

\bibitem{Jimenez2010}
A.~Jim\'{e}nez, F.~Seco, J.~Prieto, and J.~Guevara, ``Indoor pedestrian
  navigation using an {INS/EKF} framework for yaw drift reduction and a
  foot-mounted {IMU},'' in {\em Positioning Navigation and Communication
  (WPNC), 2010 7th Workshop on}, (Dresden, Germany), 11-12 Mar. 2010.

\bibitem{Farrell2008}
J.~A. Farrell, {\em Aided Navigation}.
\newblock Mc Graw Hill, 2008.

\bibitem{Bageshwar2009}
V.~L. Bageshwar, D.~Gebre-Egziabher, W.~L. Garrard, and T.~T. Georgiou,
  ``Stochastic observability test for discrete-time kalman filters,'' {\em
  Journal of Guidance, Control, and Dynamics}, vol.~32, no.~4, pp.~1356--1370,
  2009.

\bibitem{Rubia2013}
E.~de~la Rubia and A.~Diaz-Estrella, ``Improved pedestrian tracking through
  kalman covariance error selective reset,'' {\em Electronics Letters},
  vol.~49, no.~7, pp.~464--465, 2013.

\bibitem{Oberg1993}
T.~\"{O}berg, A.~Karsznia, and K.~\"{O}berg, ``Basic gait parameters: reference
  data for normal subjects, 10-79 years of age,'' {\em Journal of
  Rehabilitation Research}, vol.~30, no.~2, pp.~210--223, 1993.

\bibitem{Simon2010}
D.~Simon, ``Kalman filtering with state constraints: a survey of linear and
  nonlinear algorithms,'' {\em Control Theory Applications, IET}, vol.~4,
  pp.~1303 --1318, august 2010.

\bibitem{Huber2008}
M.~F. Huber and U.~D. Hanebeck, ``Gaussian filter based on deterministic
  sampling for high quality nonlinear estimation,'' in {\em Proceedings of the
  17th IFAC World Congress (IFAC 2008)}, (Coex, South Korea), 6-11 Jul. 2008.

\bibitem{Kurazume1994}
R.~Kurazume, S.~Nagata, and S.~Hirose, ``Cooperative positioning with multiple
  robots,'' in {\em Robotics and Automation, Proc. 1994 IEEE International
  Conference on}, vol.~2, pp.~1250--1257, 1994.

\bibitem{DeAngelis12}
A.~{De Angelis}, P.~H{\"a}ndel, and J.~Rantakokko, ``Measurement report: Laser
  total station campaign in {KTH R1 for Ubisense} system accuracy evaluation,''
  tech. rep., KTH Royal Institute of Technology, 2012.
\newblock QC 20120618.

\bibitem{Nilsson2013}
J.-O. Nilsson, C.~Sch\"{u}ldt, and P.~H\"{a}ndel, ``Voice radio communication,
  3{D} audio and the tactical use of pedestrian localization,'' in {\em Indoor
  Positioning and Indoor Navigation (IPIN), 2013 International Conference on},
  (Montb\'{e}liard - Belfort, France), 28-31 Oct. 2013.

\end{thebibliography}


\ifthenelse{\boolean{publ}}{\end{multicols}}{}

\end{bmcformat}
\fi

\end{document}